\documentclass[fleqn,10pt]{wlscirep}
\usepackage[utf8]{inputenc}
\usepackage[T1]{fontenc}
\usepackage{lineno}
\usepackage{siunitx}
\usepackage{url}
\usepackage{graphicx}

\usepackage{float}
\usepackage[numbers]{natbib}
\usepackage{tabularx}
\usepackage{arydshln}
\usepackage{caption}
\usepackage{subfigure}
\usepackage{textcomp}
\usepackage{manyfoot}%
\usepackage{booktabs}%
\usepackage{algorithm}%
\usepackage{algorithmicx}%
\usepackage{algpseudocode}%
\usepackage{listings}%
\usepackage{lipsum}
\usepackage{ulem}
\usepackage{wrapfig}
\usepackage{enumitem}
\usepackage{indentfirst}
\usepackage{todonotes}
\usepackage[font=small,labelfont=bf]{caption} % Optional: For better caption formatting
\usepackage{url}
\usepackage{mdframed}

\definecolor{codegreen}{rgb}{0,0.6,0}
\definecolor{codegray}{rgb}{0.5,0.5,0.5}
\definecolor{codepurple}{rgb}{0.58,0,0.82}
\definecolor{backcolour}{rgb}{0.95,0.95,0.92}

\definecolor{delim}{RGB}{20,105,176}
\definecolor{numb}{RGB}{106, 109, 32}
\definecolor{string}{rgb}{0.64,0.08,0.08}

\title{WxC-Bench: A Novel Dataset for Weather and Climate Downstream Tasks}

\author[1,*, $\dag$]{Rajat Shinde}
\author[1,$\dag$]{Christopher E. Phillips}
\author[1, $\dag$]{Kumar Ankur}
\author[2, $\dag$]{Aman Gupta}
\author[4, $\dag$]{Simon Pfreundschuh}
\author[1,*, $\dag$]{Sujit Roy}
\author[1]{Sheyenne Kirkland}
\author[1]{Vishal Gaur}
\author[1]{Amy Lin}
\author[2]{Aditi Sheshadri}
\author[1]{Udaysankar Nair}
\author[3]{Manil Maskey}
\author[3]{Rahul Ramachandran}
\affil[1]{Earth System Science Center, The University of Alabama in Huntsville, Huntsville, AL, USA}
\affil[2]{Stanford University, Stanford, CA, USA}
\affil[3]{NASA Marshall Space Flight Center, Huntsville, AL, USA}
\affil[4]{Department of Atmospheric Science, Colorado State University, Fort Collins, CO, USA}

\affil[*]{corresponding author(s): Rajat Shinde (rajat.shinde@uah.edu), Sujit Roy (sujit.roy@nasa.gov) }

\affil[$\dag$]{these authors contributed equally to this work}

\begin{abstract}
High-quality machine learning (ML)-ready datasets play a foundational role in developing new artificial intelligence (AI) models or fine-tuning existing models for scientific applications such as weather and climate analysis. Unfortunately, despite the growing development of new deep learning models for weather and climate, there is a scarcity of curated, pre-processed machine learning (ML)-ready datasets.  Curating such high-quality datasets for developing new models is challenging particularly because the modality of the input data varies significantly for different downstream tasks addressing different atmospheric scales (spatial and temporal).
Here we introduce WxC-Bench (Weather and Climate Bench), a multi-modal dataset designed to support the development of generalizable AI models for downstream use-cases in weather and climate research. WxC-Bench is designed as a dataset of datasets for developing ML-models for a complex weather and climate system, addressing selected downstream tasks as machine learning phenomenon. WxC-Bench encompasses several atmospheric processes from meso-$\beta$ (20 - 200 km) scale to synoptic scales (2500 km), such as aviation turbulence, hurricane intensity and track monitoring, weather analog search, gravity wave parameterization, and natural language report generation. We provide a comprehensive description of the dataset and also present a technical validation for baseline analysis. The dataset and code to prepare the ML-ready data have been made publicly available on Hugging Face - \href{https://huggingface.co/datasets/nasa-impact/WxC-Bench}{https://huggingface.co/datasets/nasa-impact/WxC-Bench}.

\end{abstract}

\begin{document}

\maketitle

\thispagestyle{empty}

\section*{Background \& Summary}

\begin{figure}[H]
    \centering
    \includegraphics[width=1\linewidth]{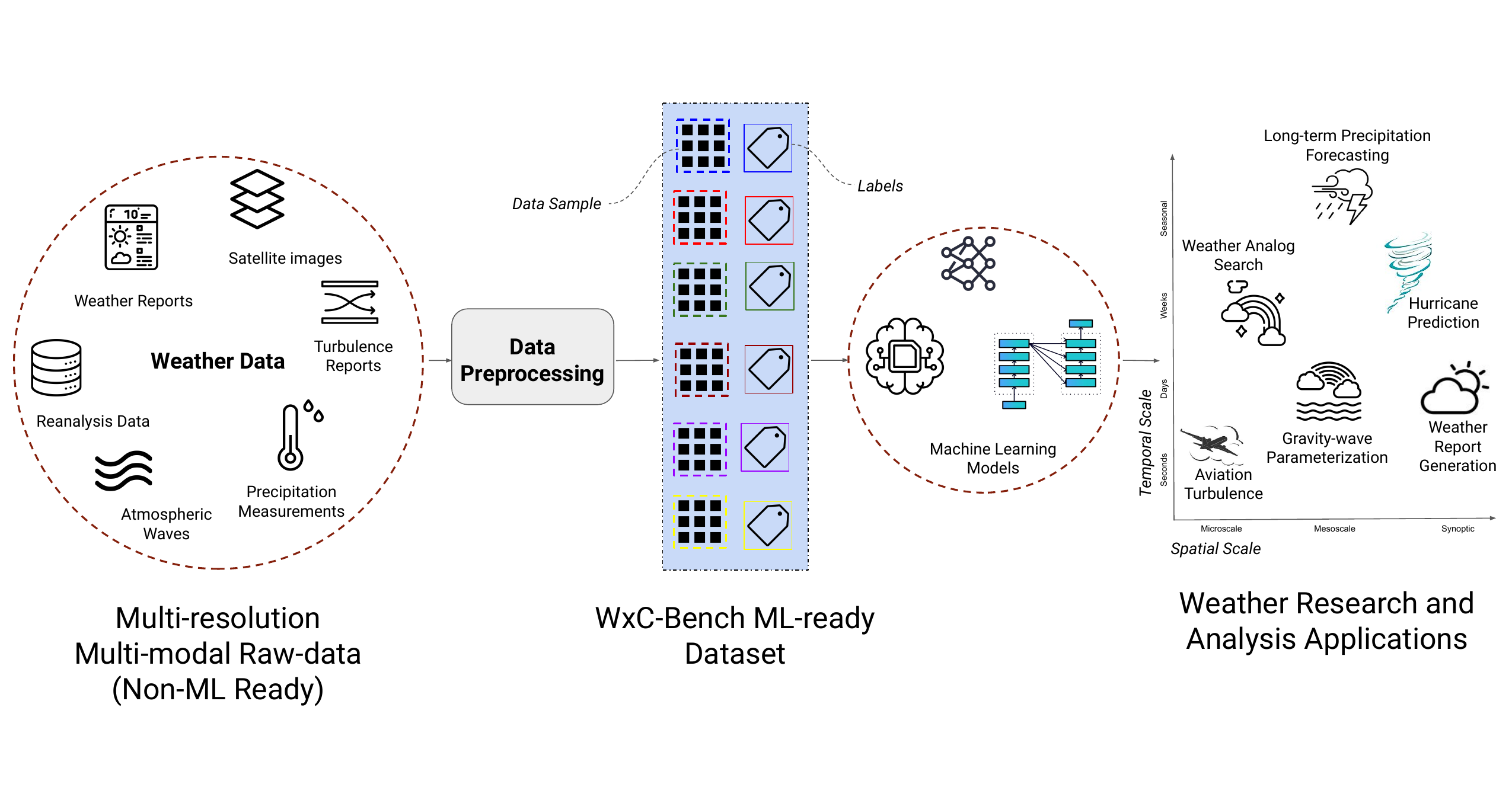}
    \caption{Illustration showing the applicability of the proposed WxC-Bench in the overall AI/ML workflows after converting raw non-ML ready datasets to ML ready datasets. The ML-ready datasets can then be directly used for training new AI models or fine-tuning pre-trained AI models. The WxC-Bench dataset can be used with AI models relevant to downstream applications across multiple spatial and temporal scales and resolutions.}
    \label{fig:overall_contribution}
\end{figure}

Weather and climate variability influences all spheres of human activity including the economy, transport, and energy production. Extreme events such as hurricanes, wildfires, droughts, and floods often result in catastrophic damage to human lives and capital. Since 1980, the U.S. alone has incurred damages worth more than \$2.5 trillion from weather and climate disasters \citep{smith2022us}. 
Current forecast models can reliably predict atmospheric extremes only up to 14-15 days in advance with forecast skill decreasing with time\footnote{{\url{https://www.whitehouse.gov/wp-content/uploads/2023/04/PCAST\_Extreme-Weather-Report\_April2023.pdf}}}. In the past, such limits of predictability were believed to be improved mostly by recording more comprehensive set of observations, and through improving traditional models. However, the past couple of years have witnessed the emergence of a radically new, AI-driven approach. 

Artificial Intelligence driven weather forecasting models deliver accurate weather predictions more quickly and cost-effectively than traditional numerical models. Notable examples of these models are FourCastNet \citep{kurth2023fourcastnet}, Graphcast \citep{lam2022graphcast}, FengWu \citep{chen2023fengwu}, Pangu \citep{bi2022pangu}, and MetNet \citep{sonderby2020metnet}. Machine Learning (ML) models also hold potential for longer-term subseasonal-to-seasonal modeling through efficient ensemble generation or otherwise \cite{Weyn.etal2021}. Additionally, ML-based methods are being explored to improve climate prediction \cite{Mansfield.etal2023}, development of ML-driven parameterizations \cite{Rasp.etal2018,Zanna.Bolton2020,Espinosa.etal2021,Wang.etal2022}, bias correction \cite{Bretherton.etal2022}, and understanding climate change impacts \cite{Davenport.Diffenbaugh2021, Diffenbaugh.Barnes2023}. Recently, AI-Numerical Weather Prediction (NWP) models, primarily using transformer-based architectures, blended with Fourier Neural Operators \citep{kurth2023fourcastnet}, Clifford Neural Operators \citep{royclifford} (NOs), and Graph Neural Networks have been introduced. These techniques aim to learn complex, nonlinear mappings required for approximating Partial Differential Equations (PDEs) that capture the inherent atmospheric physics.

Unlike traditional NWP models, which are widely used to study weather extremes, ozone recovery, air quality monitoring, parameterization development, boundary layer turbulence, etc., these pioneering models are typically trained on large volumes of data, but only to perform one specific task: medium-range weather forecasting.
This can severely limit their broader applicability, explaining the rising push to develop general-purpose AI models.

Recent models such as Prithvi WxC \citep{schmude2024prithviwxcfoundationmodel}, ClimaX \citep{climax}, Aurora \citep{bodnar2024aurora}, and Prithvi \citep{jakubik2023foundation} have been developed as a generalist AI-system capable of performing multiple tasks. These generalist models are termed foundation models (FMs) for earth sciences. FMs \citep{bommasani2021opportunities}, which were first proposed in the fields of natural language processing (NLP) and computer vision (CV), are general-purpose AI models that replace specialized task-specific models across a wide range of applications. FMs are trained in an unsupervised manner on a large volume of datasets. Within atmospheric sciences, Prithvi WxC FM \citep{schmude2024prithviwxcfoundationmodel}, a 2.3 billion parameter model has been designed for weather and climate research with a focus on applying it to multiple downstream tasks.
 
The emergence of weather and climate FMs like Prithvi WxC, ClimaX, etc., also poses questions about their generalization over several scientific applications and inherent uncertainties. Therefore, to empower a systematic benchmarking and performance evaluation of such models, there is a need for a unified ML ready dataset that spans over multiple spatial and temporal scales. WxC-Bench, the ML-ready dataset presented in this study, empowers such evaluations by facilitating the application of FMs to support downstream tasks and allowing a scoring of their performance.

In this context, Table \ref{tab:table1} outlines a few prominent ML-ready datasets that already exist. These include WeatherBench \citep{rasp2020weatherbench} and WeatherBench2 \citep{rasp2024weatherbench} exclusively for consistently scoring AI forecasts. Another dataset, ClimateNet \citep{prabhat2020climatenet}, provides a collection of expert-labeled tropical cyclones and atmospheric rivers. On longer timescales, ClimateBench \citep{watson2022climatebench} and ClimSim \citep{yu2024climsim} both provide curated climate model datasets for training climate emulators. Additionally, data processing tools for processing data into diverse ML tasks, such as  Climatelearn \citep{nguyen2024climatelearn} have been developed. The main contribution of this paper is an extension of our previous work \citep{shinde2024windset} with detailed analysis and comprehensive evaluation of the datasets. 

\begin{table*}
    \centering
    \caption{Recently proposed weather and climate data benches and their key features}
    \begin{tabular}
{|p{0.2\columnwidth}|p{0.1\columnwidth}|p{0.6\columnwidth}|}
    \hline
        Paper Title & Year of Publication & Key Features \\ 
    \hline
         WeatherBench \cite{rasp2020weatherbench}, WeatherBench 2 \cite{rasp2024weatherbench} & 2020, 2024  &
        Dataset for data-driven medium-range weather forecasting (specifically 3–5 days). Dataset curated from ERA5 re-gridded to lower resolutions (5.625° (32 × 64 grid points), 2.8125° (64 × 128 grid points), and 1.40525° (128 × 256 grid points) to be used for deep learning approaches.
         \\
         ClimateNet \cite{prabhat2020climatenet} & 2020 & Dataset trained on expert-labeled climate data via ClimateContours, enables precise weather pattern segmentation, which utilizes output data from the 25 km Community Atmospheric Model (CAM5.1). This is a curated expert-labeled dataset which is trained on the DL segmentation model.  \\
         ClimateBench \cite{watson2022climatebench} & 2022  & Dataset curated from a suite of Earth System Models participating in the Coupled Model Intercomparison Project 6 (CMIP6), covering a variety of global warming scenarios as yearly averages, at the native grid resolutions ($\sim$2$^{\circ}$).\\
         ClimSim \cite{yu2024climsim} & 2024  & Dataset compiled using the E3SM-MMF multi-scale climate simulator and stored every 20 minutes for 10 years, at an approximate grid-spacing of 1.5$^{\circ}$ using a cubed-sphere unstructured grid. Curated for hybrid physics-based ML climate emulation.\\
         Climatelearn \cite{nguyen2024climatelearn} & 2024 & Dataset curated from the ERA5, extreme ERA5, Coupled Model Intercomparison Project Phase 6 (CMIP6) and PRISM, and has been used for weather forecasting, downscaling and climate projection. \\
    \hline
    \end{tabular}
    \label{tab:table1}
\end{table*}

Expanding the scope of the tasks and data on which the ML model's transfer learning capabilities are evaluated is important for weather and climate applications given spatial and temporal complexities. To explore and assess the potential of generalizable AI models / FMs for meteorology research, it is imperative that the considered tasks  (a) span multiple spatial and temporal scales, from forecasting, to seasonal prediction, to climate scale prediction (Figure \ref{fig:overall_contribution} right), and (b) vary in input formats and output applications that require assimilating data from multiple sources. 

The main contribution of this paper is WxC-Bench (See Figure \ref{fig:overall_contribution}, Appendix A),  a multi-modal dataset that supports not only the evaluation of generalizable AI models for weather and climate research but also in development of state-of-the-art ML benchmarks for scientific tasks. The dataset can be used for tasks spanning multiple spatio-temporal scales (Figure \ref{fig:overall_contribution} right). While creating WxC-Bench, the tasks are chosen to sample a variety of ML applications, from natural language generation to image classification to regression, and span atmospheric processes ranging from meso-$\beta$ (20 - 200 km) to synoptic scales (2500 km).\\

\section*{Downstream Task Descriptions and Datasets}

This paper presents a compilation of 6 ML-ready datasets for developing task-specific ML models. The tasks are selected to represent a broad scale of the atmospheric processes and challenges (Table \ref{tab:atmospheric_phenomena}) while also having useful applications for research and day-to-day human activities (See Relevance column of Table \ref{tab:atmospheric_phenomena}). Some tasks, such as gravity wave (GW) parameterization, are considered relatively straightforward, while others like natural language forecasting are much more complicated attributing to learning from multi-modal features. This enables a comprehensive and complete assessment of a trained AI model's transfer learning capability for generalization across atmospheric science.

\begin{table*}[htbp]
    \centering
    \small
    \caption{Comparison of Downstream Tasks as Machine Learning (ML) Phenomenon used in this study}
    \label{tab:atmospheric_phenomena}
    \begin{tabular}{|p{2cm}|p{3cm}|p{3cm}|p{3cm}|p{5cm}|}
        \hline
        \textbf{ML Phenomenon} & \textbf{Downstream Task} & \textbf{Temporal Scale} & \textbf{Spatial Scale} & \textbf{Relevance} \\
        
        \hline
        Classification & Aviation Turbulence & Upto seconds & Microscale & Vital for flight safety and predicting turbulent regions \\
        \hline
        Parameterization & Atmospheric Gravity Waves & Upto Day & Mesoscale &  Crucial for understanding momentum transfer and vertical couplings in the atmosphere \\
        \hline
        Search & Weather Analog Search & Hours & Meso-to-Synoptic scale &  Helps in finding past weather patterns similar to current conditions for forecasting \\
        \hline
        Prediction & Long Term Precipitation Forecasting & Subseasonal & Meso-to-Synoptic scale & Essential for water resource management and agricultural planning \\
        \hline
        Prediction & Hurricane & Weeks & Mesoscale & Critical for predicting hurricane paths and impacts, saving lives and property \\
        \hline
        Multi-modal generation & Natural Language Forecasting & Daily & Synoptic & Communicates complex meteorological data into understandable forecasts for day-to-day activities \\
        \hline
    \end{tabular}
\end{table*}

In the following sections, we elaborate on the dataset for the above-mentioned downstream tasks representing an atmospheric phenomenon. 

\subsection*{Aviation Turbulence Prediction}

Aviation Turbulence in the lower and middle atmosphere presents risks for passenger and cargo airliners, especially when it is encountered unexpectedly \cite{10.1029/2020gl086999,
10.15394/jaaer.2002.1301}. Turbulence, however, is a microscale feature that defies direct prediction by numerical weather prediction models with even relatively fine grid-spacing on the order of 1 Km. While machine learning techniques have been explored for turbulence prediction based on information from numerical weather prediction or onboard sensors \cite{MLturbulence2021,10.1007/s10994-013-5346-7,10.1016/j.mlwa.2020.100008}, deep learning (DL) techniques remain  under-investigated. Thus, there is a scope for sophisticated data-driven approaches to improve turbulence forecasts for aviation, which can be supported by this dataset.

\paragraph*{Dataset Description}

The training dataset prepared for aviation turbulence detection is derived from the Modern-Era Retrospective Analysis for Research and Applications, Version 2 (MERRA-2) re-analysis data set \cite{10.1175/jcli-d-16-0758.1}. MERRA-2 is available on a 0.625$^{\circ}$$\times$0.5$^{\circ}$ grid resolution with 72 hybrid model vertical levels and 42 output pressure levels. With this grid spacing, MERRA-2 resolves synoptic scale features such as frontal zones and midlatitude cyclones, which can impact turbulence production but not fine-scaled processes such as boundary layer growth. MERRA2 is used to provide temperature, relative humidity, zonal and meridional winds, vertical motion, orography, and geopotential height. These variables are retrieved for the lowermost 40 pressure levels, spanning from the surface to 100 hPa. Each input feature is valid at 1800 UTC, corresponding to approximately local noon (Central Time) over the contiguous United States (CONUS).

\begin{figure}[H]
    \centering
    \includegraphics[width=0.7\linewidth]{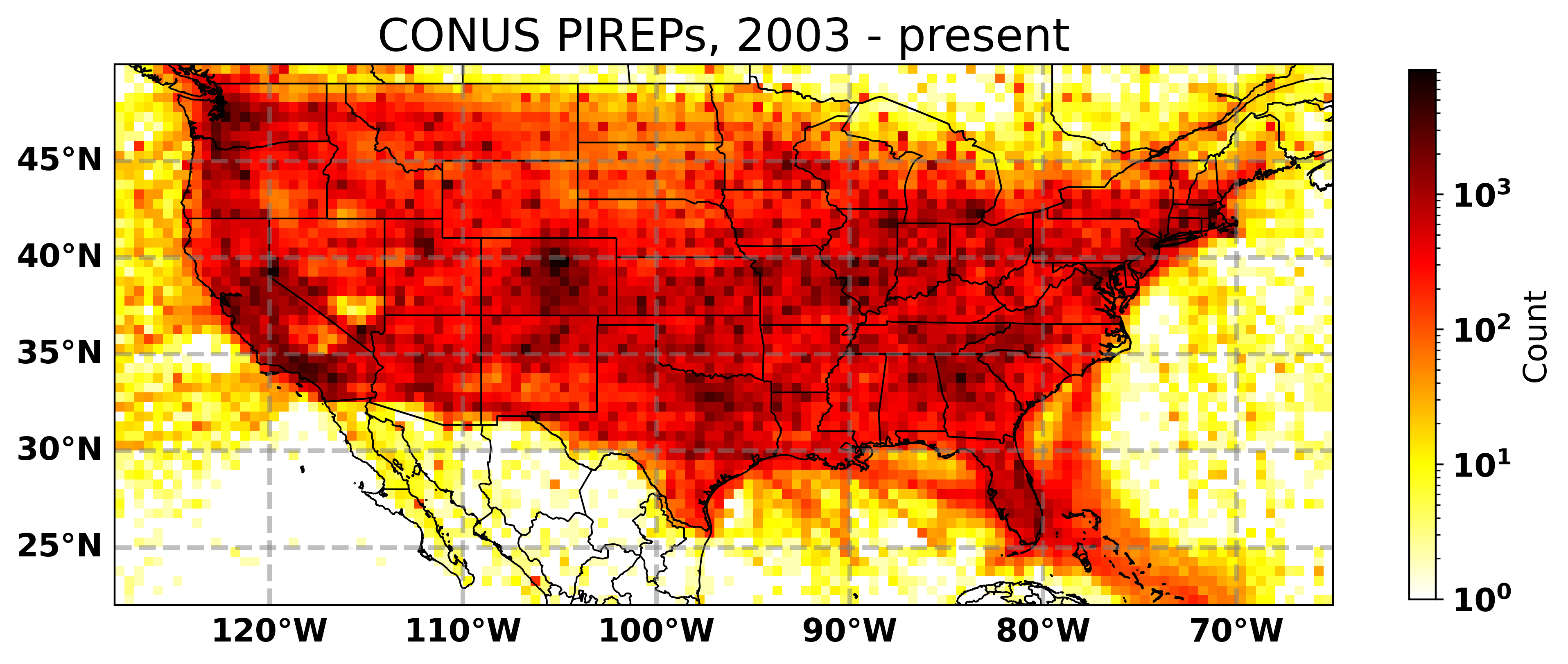}
    \caption{Spatial distribution of PIREPs turbulence reports over CONUS from 2003-Present. Note that values are scaled logarithmically.}
    \label{fig:pireps_map}
\end{figure}

The training labels for this task are Pilot Reports (PIREPs). PIREPs are point-location reports written by pilots that describe in-flight conditions and can include a variety of information, such as cloud layers, general weather conditions, icing, turbulence encountered, and more. The PIREPs are retreived via the Iowa State University archives \citep{iowastate} and have a temporal extent since January 1, 2003, to the present day. Currently, data used span from January 1, 2003, to early November 6, 2023. The PIREPs recorded over the CONUS are primarily considered for this dataset. The spatial distribution of the turbulence reports are displayed in Figure \ref{fig:pireps_map}.

The PIREP reports were first grouped by flight levels into bins of 0-14, 900 ft (LOW), 15,000-29,900 ft (MIDDLE), and 30,000 ft or greater (HIGH). These levels were selected as they correspond approximately to the planetary boundary layer (PBL), free atmosphere, and tropopause respectively. Flight levels above 50,000 ft were removed as this is typically above the flight level of commercial aircraft. After filtering flight levels, there were 6,004,516 total data points.

\begin{table}[ht]
\centering
\caption{Distribution of Turbulence Across Levels}
\begin{tabular}{lcccc}
\hline
Level   & Turbulent & Non-turbulent & \# Samples \\ \hline
LOW     & 7\%       & 93\%          & 15910      \\
MIDDLE  & 6\%       & 94\%          & 8486       \\
HIGH    & 7\%       & 93\%          & 13551      \\ \hline
\end{tabular}
% \caption{Distribution of Turbulence Across Levels}
\label{tab:turbnet_classes}
\end{table}

To convert turbulence reports into binary classification labels, all reports are re-classified as moderate or greater (MODG) following prior studies \cite{10.1175/2008jamc1799.1, 10.1016/j.mlwa.2020.100008}. PIREPs are then binned into the nearest MERRA-2 grid cell by day, and any cell with MODG report frequency more than 25\% is considered "turbulent" for that day. Initially, the reports were binned on 3-hourly intervals to match MERRA-2 output, but this was found to have very few reports within each cell for each time period. The distribution of labels for each level is shown in Table \ref{tab:turbnet_classes}. 

%The training data are technically validated in section \ref{sec:AVTvalidation}.

\subsection*{Gravity Wave (GW) Parameterization}

Atmospheric mesoscale processes like clouds, gravity waves (GWs), and precipitation, have global impacts on the atmospheric flow but are often  \textit{parameterized} in weather and climate models \cite{Fritts.Alexander2003, Kim.etal2003, Achatz.etal2023}. Climate model parameterizations are typically designed to be strictly vertical i.e., single column \citep{Sorbjan2009, Chen.etal2018, Plougonven.etal2020}, due to which they neglect the horizontal evolution of mesoscales. Particularly in the context of meso-$\alpha$ and meso-$\beta$ GWs, this creates a major representation bias in weather and climate models, leading to global momentum imbalances such as colder-than-observed temperatures in the high latitudes (termed as cold-pole bias) or biases in global mass transport \citep{McLandress.etal2012, delaCamara.etal2018, Gupta.etal2024}. This means that the representation of horizontal propagation of atmospheric GWs, and in general, other parameterized mesos-$\alpha$ and meso-$\beta$ atmospheric processes, is crucial for both regional short-term weather forecasting and global long-term climate prediction \cite{Kruse.etal2022, Kim.etal2023,Gupta.etal2024a}.

GWs couple the different layers of the atmosphere and demonstrate significant horizontal motion as they propagate. This property makes them an excellent candidate for simulation using a specialized AI-model or a generalized FM, and for a test of the AI's capability to learn regional variability and vertical coupling of the atmosphere. A successful ML-based prediction of subgrid-scale GW momentum fluxes opens avenues to first learn missing model physics from high-resolution observations and then represent the missing mesoscale physics in coarse-climate models using ML. The strategy can be applied to various other mesoscale processes, which are currently parameterized using (approximate) single-column schemes in weather and climate models. Availability of curated datasets, such as this, thus promotes the development of fast and potentially more physically accurate ML parameterizations of atmospheric processes. 

\begin{figure}[H]
    \centering
    \includegraphics[scale=0.6]{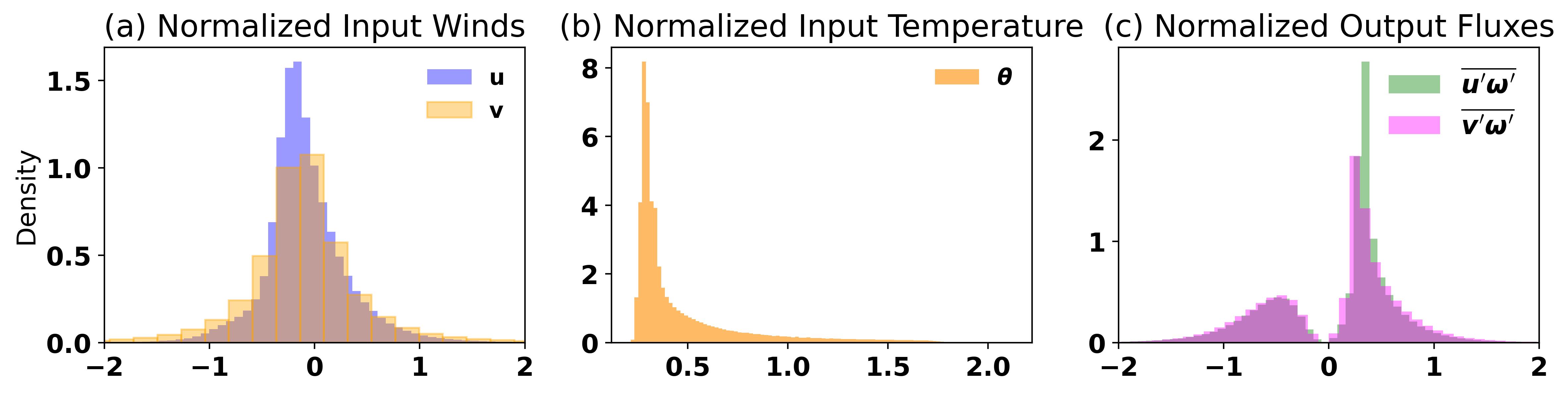}
    \caption{Distribution of the coarse-grained input features (scaled background atmospheric state) and output labels (scaled GW momentum fluxes) for a single month (January 2010) from ERA5.}
    \label{fig:gw_dist}
\end{figure}

%\paragraph{Data description}
\paragraph*{Data Description}

For a given background state of the atmosphere, GW parameterizations in climate models predict the associated subgrid-scale momentum fluxes due to GWs \citep{Kim.etal2003}. Here, this problem is formulated as a regression task to predict the resolved GW fluxes associated with an input background state.

The subgrid-scale momentum fluxes, i.e., the output, are computed using Helmholtz decomposition. Helmholtz decomposition of the full horizontal velocity field $\vec{u}=(u,v)$ decomposes the full flow into its rotational (balanced) and divergent (unbalanced) parts, i.e., $(u,v) = \vec{u} = \vec{u}_{rot} + \vec{u}_{div}$. The divergent part is associated with GWs \citep{Lindborg2015}. A high-pass filter is then applied to remove the first 21 harmonics of the divergent flow. The eddy vertical velocity, $\omega'$, is computed by removing the zonal mean from full $\omega$. The divergent part $\vec{u}_{div}$ and $\omega'$ are then used to
compute the fluxes, ($F_u$, $F_v$) = ($\overline{u_{div}\omega'}$, $\overline{v_{div}\omega'}$ ).

The training data for global gravity wave parameterizations is prepared using the modern reanalysis, ECMWF Reanalysis v5 (ERA5), from the European Centre for Medium-Range Weather Forecasting (ECWMF) \citep{Hersbach.etal2020}. ERA5 is publicly available at a horizontal resolution of 0.3$^{\circ}$$\times$0.3$^{\circ}$ and 137 hybrid pressure levels at an hourly temporal resolution. Accounting for grid-scale hyper-diffusion and other numerical effects, ERA5 resolves GWs with wavelengths exceeding 150-200 km and longer \citep{Skamarock2004, Gupta.etal2021a, Pahlavan.etal2021}. The global wind and temperature data are retrieved on 122 vertical levels: from 1.45 hPa (model level 16) to the surface (model level 137), and are used to compute the training labels, i.e., the momentum fluxes due to GWs. The top 15 levels near the model top are removed to avoid the artificial sponge damping in the mesosphere.

The background winds and temperature at ERA5's native grid resolution are conservatively coarse-grained to a 2.8$^{\circ}$$\times$2.8$^{\circ}$ (64$\times$128) Gaussian grid to obtain the background state of the atmosphere, i.e., the training input. Likewise, the output fluxes are computed on the native grid and then conservatively coarse-grained to the 2.8$^{\circ}$ grid, as denoted by the overbar. At this resolution, WxC-Bench provides a detailed, multi-year, global, evolution of subgrid-scale fluxes, far superseding other existing datasets, e.g., ClimSim \cite{yu2024climsim}. Atmospheric GWs exist on spatial scales ranging from O(1) km to O(1000) km. We do not claim that ERA5, at 30 km horizontal resolution, completely resolves these scales. However, given the dearth of storm-resolving global climate simulations, ERA5 still presents as one of the most extensive long-term records to study atmospheric GWs with wavelengths $\sim$150 km and above.

% Input features
The input feature vector consists of three dynamical variables:  zonal wind ($u$), meridional wind ($v$), and potential temperature, $\theta = T(p/p_0)^{\gamma}$, where $T$ is the temperature, $p$ is the pressure, $p_0=1000$ hPa is a reference surface pressure, and $\gamma=-0.286$ is a constant). Each quantity is a function of longitude, latitude, pressure (height), and time. Note that the input comprises the full winds, not just the divergent component. The output contains the vertical flux of zonal and meridional momentum carried by GWs, i.e., {$u'\omega', v'\omega'$}. Similar to the input, the output is computed globally for each longitude, latitude, pressure (height), and time. For this downstream task, input features and labels are computed using hourly-ERA5 for four years, viz., 2010, 2012, 2014, and 2015. These years were chosen based on rich wintertime orographic GW activity in the Southern Hemisphere. Thus, the dataset comprises of a total of 64$\times$128$\times$24$\times$1461 ($\sim287$ million) columns of training and validation data.

Additionally, the input and output are scaled differently, i.e., the winds, temperature, etc., are normalized using their respective 4-year global mean and standard deviation. Subsequently, the means and deviations are used to scale the input variables as follows:

\begin{equation}
    u_{scaled}(t, p, \phi, \lambda) = \frac{u'(t, p, \phi, \lambda)}{3\sigma_u} = \frac{u(t, p, \phi, \lambda) \;-\; \mu_u}{3\sigma_u}
\end{equation}
where $\mu_{u}$ is the temporally and zonally averaged zonal wind and $\sigma_{u}$ is the standard deviation. Identical normalization is performed for $v$. The scaling transforms the input values to a range within $[-2,2]$. The potential temperature, $\theta$, was scaled simply by dividing by 1000. 

GW generation is highly intermittent; thus, the GW fluxes can have a symmetric exponential (Laplace) distribution with small values around zero. Therefore, the GW fluxes are scaled by their respective global means and standard deviations and then a cube root is applied, which transforms very small and very large values and maps them closer to 1. The normalized variables are then concatenated to form the input and output variables. Figure \ref{fig:gw_dist} illustrates the input and output variables distribution.

\subsection*{Weather Analog Search Dataset} 
\label{weather_analog}

Weather analogs, that is, weather events and structures that are similar to each other, serve several purposes in the Earth science community. For forecasters, past events provide insight into how current weather can evolve and serve as initial conditions for ensemble model systems \citep{chattopadhyay2020analog, van1989new}. They can serve for training deep-learning models. For researchers, weather analogs make useful case studies for analyzing how particular weather systems evolve \citep{yang2019ultra, franch2019mass}. Thus, a rapid analog search is useful for improving forecasts and aiding scientists in locating desirable historic records for their research.

\paragraph*{Data Description}

For searching analogs over a desired dataset, e.g., MERRA-2, the data samples are required to be encoded to enable rapid search over the data space. There are several ways to accomplish this such as vector search based on minimizing the distance between embedding generated by a post-tokenization module of a machine learning architecture \citep{ahn2023searching}. Also, lookup table-based searching over a database of images using transforms - Fast Fourier Transform \citep{yang2019ultra} or Wavelet “fingerprints” as search key \cite{raoult2018fast, demir2015hashing}. Searching analogs is also possible based on deep learning-based approach for generating latent features using all the temporal multivariate input predictors in this latent space rather than the original predictor space \cite{hu2023machine}.

In this context of analog search, the searching approach must be spatially constrained in some way as the global repetition of an atmospheric state is unlikely \cite{wx_analog_1989}. Thus, following scenarios emerge for potential applications from a user perspective:

\begin{itemize}
    \item \textit{A user searches for a specific phenomena, such as a mid-latitude cyclone, and is given multiple instances from across the globe.}
    
    In this scenario, the model provides users with multiple instances of mid-latitude cyclones distributed globally. Each analog is limited in area, so that matches with the user case are more likely. Such an approach holds particular utility for individuals engaged in constructing climatologies or those endeavoring to identify case days for analyzing mid-latitude cyclone-driven weather events.
    
    \item \textit{A user constrains the model to search a domain of interest.}
    
    Alternatively, the model can be tailored to search within a designated domain of interest, restricting its focus to a specific region. This configuration is particularly advantageous for forecasters with a vested interest in the weather dynamics of a particular locale and for researchers and climatologists seeking targeted insights into regional phenomena.
    
\end{itemize}

\begin{figure}[H]
    \centering
    \includegraphics[width=1\linewidth]{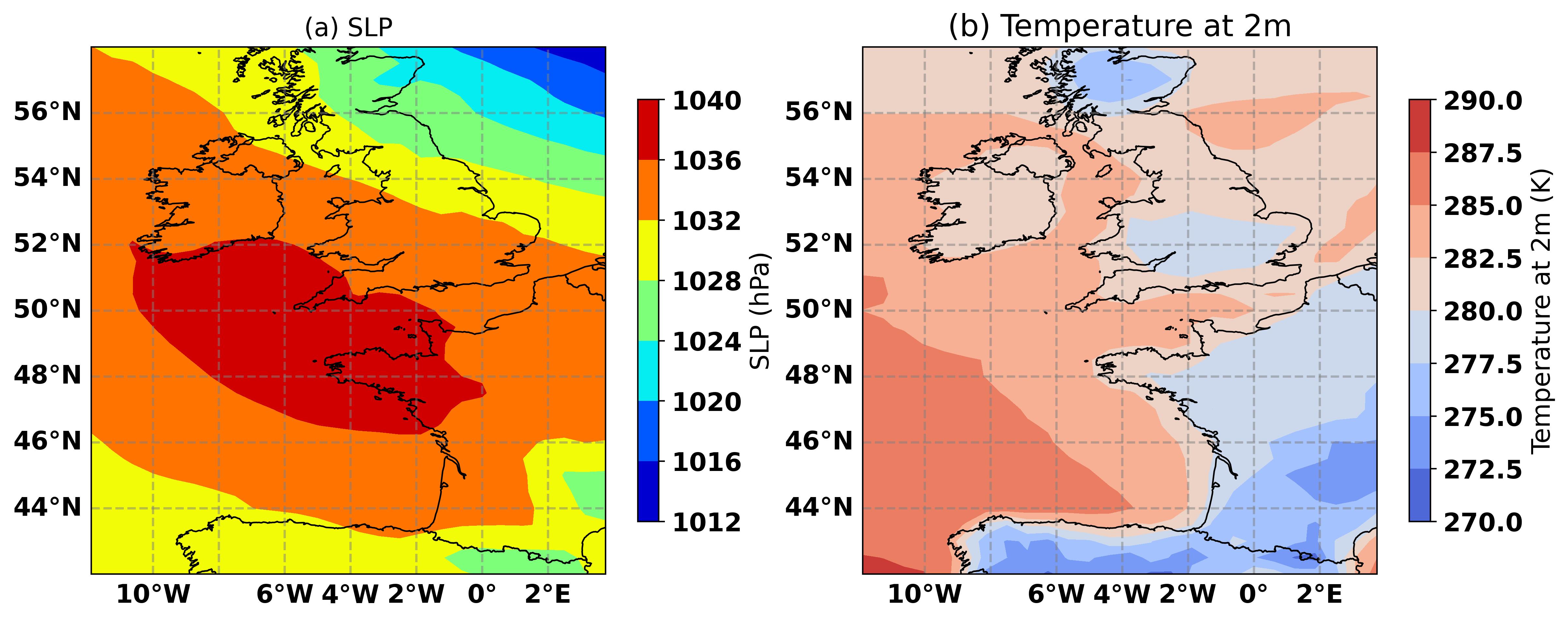}
    \caption{Illustration of training images for (a) sea level pressure (b) temperature at 2m, for January 01, 2019.}
    \label{fig:wx-analog}
\end{figure}

The above-mentioned approaches may be handled simultaneously by breaking the larger grids into sub-grids and encoding each separately with location information. Then, the search can be performed across either all or some sub-grids. The European Centre for Medium-Range Weather Forecasts (ECMWF) utilizes the sub-grid approach for its analog search tool.

Similar to the ECMWF methodology, a series of sub-grids for MERRA-2 have been created with edge lengths of approximately 1800 km (Figure~\ref{fig:wx-analog}). Note that the domain size in grid points is preserved, meaning that the edge length of the subgrids varies with latitude. Edge length of 1800 km has been shown to be adequate for forecasting features over a point of interest for at least 12 hours while maintaining a robust number of analogs available in the historical record \cite{wx_analog_1989}. Each subgrid overlaps by 10 points on the western and southern edges to mitigate issues where the point of interest is on the edge of a sub-domain. This leads to 623 subgrids for the entire globe at the MERRA-2 resolution of 0.5x0.625 degrees.

\subsection*{Long-range Precipitation Forecast}

\label{precip}

The long-term precipitation or long-range precipitation forecast task consists of predicting global, daily-accumulated precipitation four weeks into the future. This forecasts range falls within the so-called \textit{sub-seasonal to seasonal range} (S2S), which constitutes an important frontier in weather and climate research. Conventional numerical weather prediction (NWP) can produce skillful precipitation forecasts ten to twelve days into the future but beyond that their skill deteriorates to or below the level of a climatology \citep{vitart_22_s2s_challenge}. While climate models can predict weather patterns on seasonal scales, the S2S range is currently considered a predictability desert \citep{robertson_18_predictability_desert}. Since many decisions in the management of agricultural and water resources fall into this two-weeks to two-months time scale, improving precipitation forecasts several weeks into the future has significant societal value \cite{vitart_12_s2s}. The proposed dataset is targeted to address the S2S precipitation forecasting.

The input data for the long-range precipitation forecast consists of gridded, global satellite observations from a range of sensors providing observations in the visible, infrared, and microwave regions of the electromagnetic spectrum. The reference data is derived from satellite-based precipitation estimates that were corrected to match ground-based gauge measurements and arguably constitute the most reliable global estimates of precipitation currently available.

A related dataset that has been proposed is RainBench \citep{dewitt_2020_rainbench}, which combines \textit{simulated} satellite observations with satellite- and reanalysis-based precipitation estimates for machine learning-based precipitation forecasts. However, the simulated satellite observations in RainBench cover only a limited number of channels and are available only from 2016 onward, which makes the dataset less suitable for evaluating long-range forecasts of precipitation. Moreover, since the simulated observations are based on non-cloud-resolving numerical models they will inherit their limitations regarding the representation of cloud- and precipitation-related processes. Here, we propose a novel benchmark dataset for long-range precipitation forecasts based on almost forty years of satellite observations and corresponding precipitation estimates.

% \subsubsection*{Methods}

\paragraph*{Data Description}

\begin{figure}[t]
    \centering
    \includegraphics[width=1\textwidth]{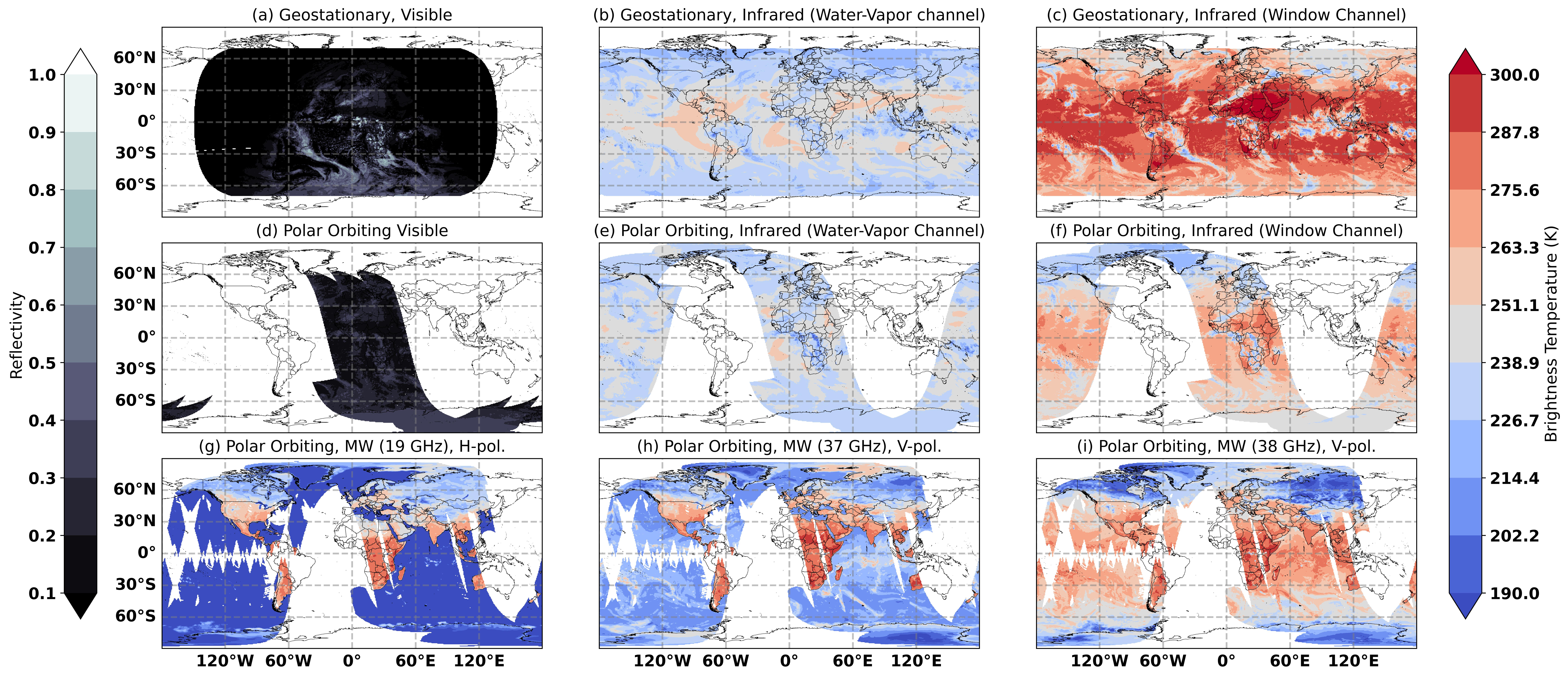}
    \caption{Example of the input data for the Long-range precipitation forecast task. Each panel displays observations from one of the channels of the three observational datasets that form the input for the Long-range precipitation forecast. Panels (a), (b), and (c) show visible and infrared geostationary observations from the GridSat-B1 dataset. Panels (d), (e), and (f) show corresponding observations from polar-orbiting satellites from the PATMOS-x dataset. Panels (g), (h), and (i) show microwave observations from the SSMI dataset.}
    \label{fig:ltpf_satellite_observations}
\end{figure}

The input data for the long-range precipitation forecast consists of gridded satellite observations from three observation sources: (1) infrared and visible observations from geostationary satellites, (2) infrared and visible observations from polar-orbiting satellites, and (3) microwave observations from polar-orbiting satellites. The input observations were chosen to maximize the temporal coverage of the training dataset. The observations are extracted from three observational records. Geostationary observations are taken from the GridSat-B1 dataset \citep{knapp_11_gridsat}, infrared and visible observations from the Advanced Very High-Resolution Radiometer (AVHRR) and the High-resolution Infra-Red Sounder (HIRS) from the PATMOS-x fundamental climate data record \citep{foster_21_patmosx}, and microwave observations from the Special Sensor Microwave/Imager (SSM/I) from the NOAA Climate Data Record (CDR) of SSMI(S) and AMSR2 Microwave Brightness Temperatures \citep{kummerow_22}. Example observations from each of these records from selected channels for 1 January 2000 are shown in Figure~\ref{fig:ltpf_satellite_observations}.

The reference precipitation estimates are derived from the PERSIANN-CDR product \citep{ashouri2015persiann} for the time range 1983 through June 2000 and the IMERG Final \citep{huffman_19_imerg} product from June 2000 through the present. In contrast to  PERSIANN-CDR, which is based solely on geostationary visible and infrared observations,  IMERG combines observations from passive microwave and geostationary sensors. Because of this, IMERG can be expected to provide better sensitivity to precipitation and is therefore used to evaluate the forecasts. However, the IMERG precipitation estimates are available only from the year 2000 onward. We therefore provide an extended training record based on the combined estimates from PERSIANN-CDR and IMERG for the case that the extended training period may prove beneficial for the training of the forecast models.

Both input and reference data are interpolated to a regular latitude-longitude grid with a zonal 
resolution of $0.625\ \unit{^\circ}$ and meridional resolution of $0.5\unit{^\circ}$. The data are organized into separate files by day and input and target source.

%A challenge with training precipitation forecasts on satellite observations is that the input-data record is not always complete.
% The biggest gap in the input data availability is due to microwave observations first becoming available in 1987 whereas data from the GridSat and PATMOS-x records is available from 1983 onwars. However, even after that, the data availability is not always complete. The ML will thus have to deal with intermittent data availability from the three input data sources.

\begin{table}[h]
\caption{
Number of daily training files for the long-range precipitation forecast task by dataset split and input data source.}
\label{tab:ltp-data}

    \begin{tabular}{ccccccc}
    \hline
    Split & Start & End & PATMOSx files & GridSat files & SSMI files & Precipitation files \\
    \hline
    Training & 1983-01-01 & 2017-12-31 & 12 591 & 11073 & 12763  & 12784 \\
    Validation & 2018-01-01 & 2018-12-31 & 358 & 358 & 358 & 358 \\
    Testing & 2019-01-01 & 2022-12-31 & 1408 & 1442 & 1442 & 1444 \\
    \hline

    \end{tabular}
\end{table}

While the dataset allows for a range of training scenarios, we propose the following evaluation protocol for the long-range precipitation forecasts task. Years 1983 through 2018 of the extracted satellite observations and matched precipitation estimates are designated as training data, while year 2019 through 1 January 2023 are used for testing. Table~\ref{tab:ltp-data} lists the number of files by input and target source for the three dataset splits. Scripts to evaluate the machine-learning-based forecasts against the conventional NWP baselines are provided together with datasets and section \ref{sec:s2sValidation} presents the technical validation of the proposed dataset.

\subsection*{Hurricane Prediction and Intensity Estimation Dataset} 
\label{hurricane}

Hurricanes are one of the most destructive natural phenomena affecting human populations, infrastructure, and economies, causing approximately \$17 billion in damages annually in just the United States. Moreover, it has been observed that the frequency of hurricane occurrence is increasing over the years \cite{emanuel2017assessing}. Further, studies show that the power dissipated by these hurricanes has doubled since the 1980s \cite{bromirski2008increasing}, likely due to rising sea surface temperatures \cite{emanuel2005increasing}. Many of these hurricanes make landfall along the coastal regions of southeast and southwest US which are zones of preferential urban growth. Thus, there is a need for accurate and timely predictions of hurricanes.

\paragraph*{Data Description}
This dataset aggregates hurricane track data from the Atlantic and Pacific basins spanning 1980 to 2022. Data is primarily sourced from the Hurricane Database (HURDAT) maintained by the United States National Hurricane Center. HURDAT was accessed through the \texttt{tropycal} package \cite{burg2021tropycal}, and the track, intensity (10m sustained wind speed and minimum sea level pressure), and size of the hurricane (radius of maximum winds and radius of 34 kt winds) were retrieved. Since the best track data is available on 6-hourly intervals, all parameters were then cubically interpolated to a 3-hourly interval to match MERRA-2 reanalysis.

\begin{figure}
    \centering
    \includegraphics[scale=0.7]{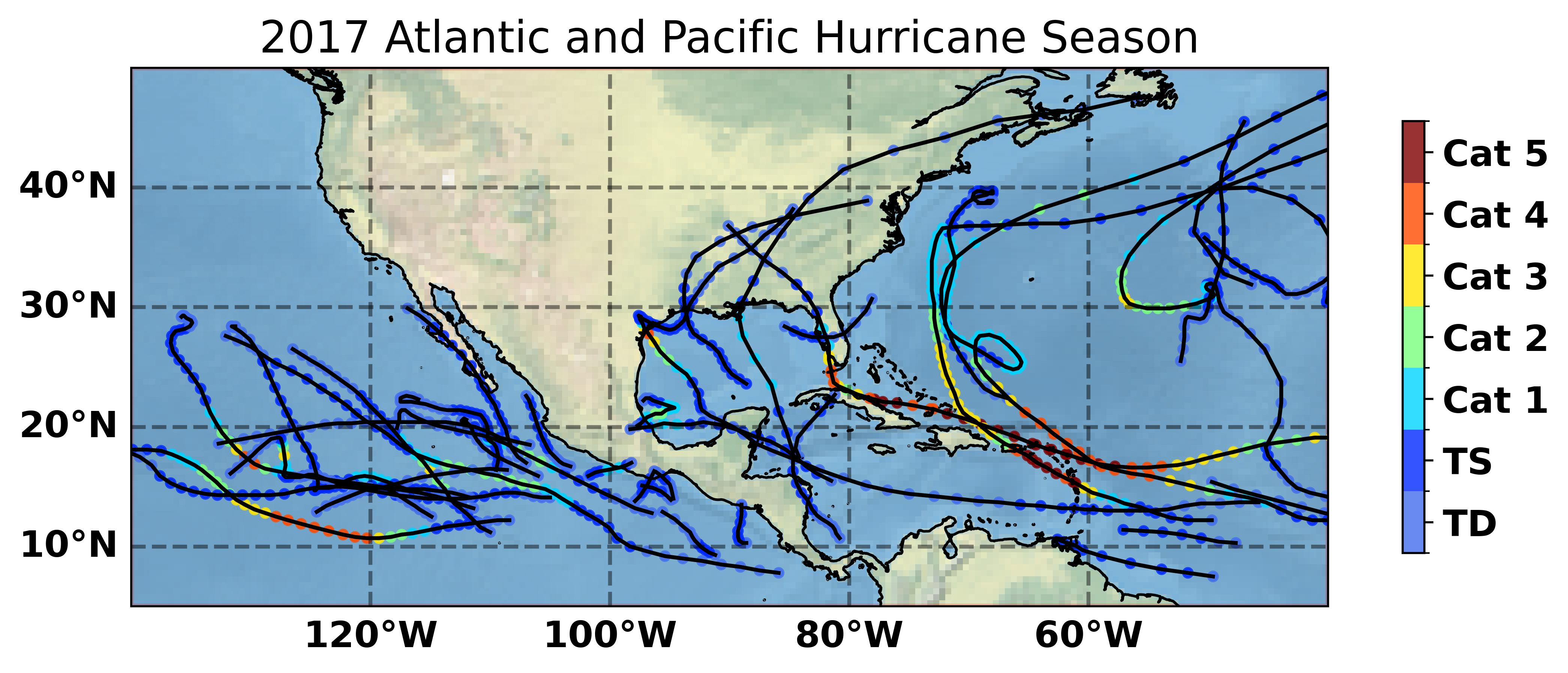}
    \caption{Tracks of named hurricanes in the Atlantic Ocean by their respective categories, ranging from tropical storms (TS) to Category 5 (Cat 5) hurricanes for the year 2017.}
    \label{fig:hurr_tracks}
\end{figure}

By compiling information on hurricanes in both the Atlantic and Pacific Ocean basins, this dataset enables models to learn from a broader, more spatially diverse dataset that can help them generalize across the globe. Figure \ref{fig:hurr_tracks} shows an example of tracks of named hurricanes in the Atlantic Ocean by their respective categories, ranging from tropical storms (TS) to Category 5 (Cat 5) hurricanes for the year 2017. This dataset encompasses a total of 649 named hurricanes, and it is noteworthy that 287 of these hurricanes fall into the TS/TD category. This category represents storms with relatively lower wind speeds but still carries the potential for significant rainfall and localized impacts. There are 120 hurricanes classified as Cat 1 hurricanes. These are characterized by wind speeds between 74 to 95 miles per hour and can damage residential and small structures. Additionally, 170 hurricanes fall into the Cat 2 and higher, signifying more intense and potentially destructive storms with higher wind speeds and the capacity to cause more extensive damage. This categorization allows for a nuanced understanding of the distribution of hurricane intensity in our dataset. It serves as a crucial reference for our study, enabling us to analyze the factors contributing to the formation and development of hurricanes across various categories.

\begin{figure}
    \centering
    \includegraphics[scale=0.6]{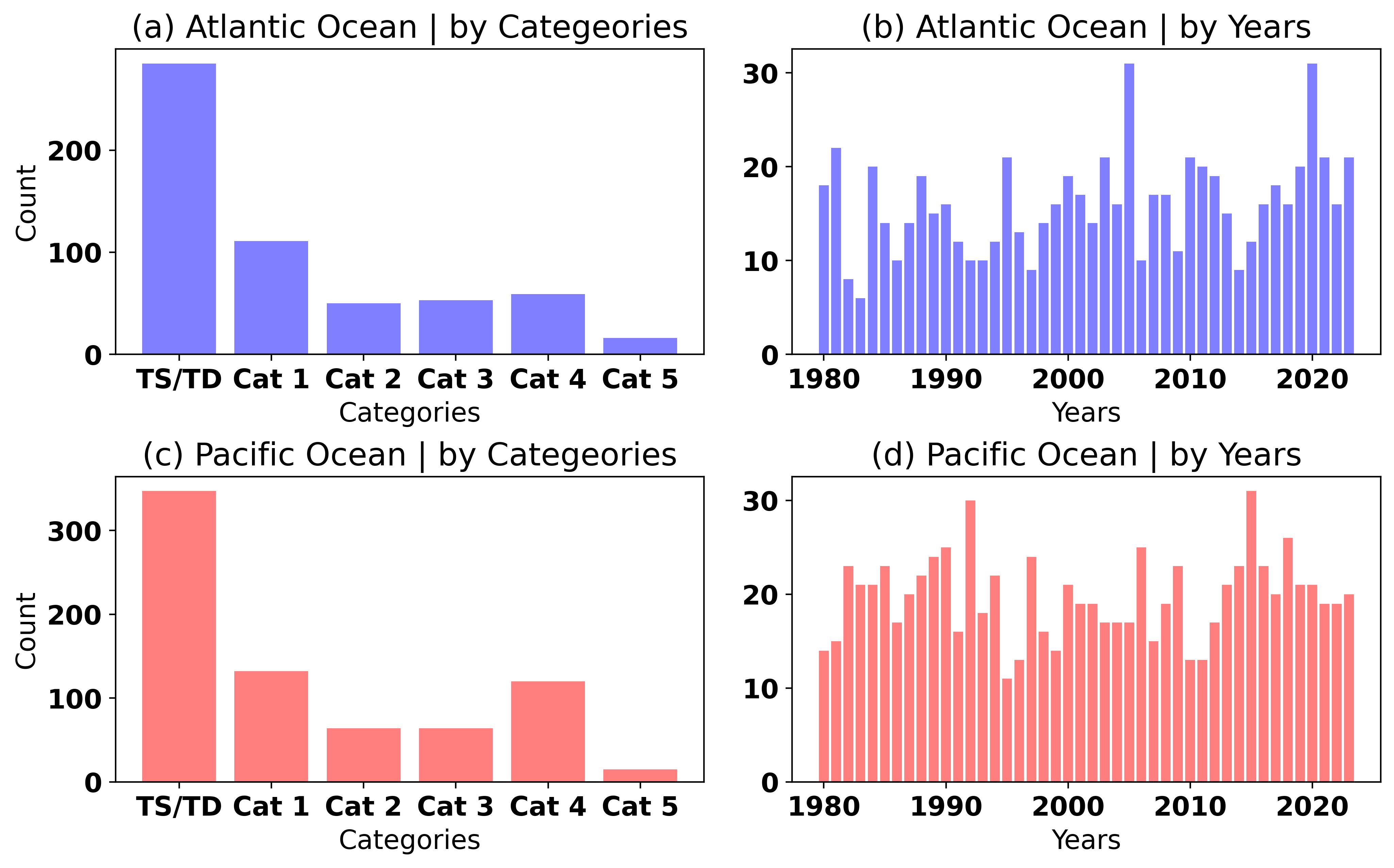}
    \caption{Hurricane counts by categories and years (1980-2022) for North Atlantic and East Pacific hurricanes.}
    \label{fig:hurr_int}
\end{figure}

Figure \ref{fig:hurr_int} (a) and (c) show the hurricanes that formed in the Pacific Ocean basin. Among the 810 hurricanes in this dataset, 350 hurricanes are in the TS/TD category. Following this, 150 hurricanes were classified as Cat 1, indicative of moderate storm intensity. There are approximately 70 hurricanes each in the Cat 2 and Cat 3, highlighting the presence of storms with significantly increased wind speeds. Furthermore, approximately 150 hurricanes are classified as Cat 4, and around 20 hurricanes fall into the most severe category, Cat 5. The Pacific Ocean basin experiences more hurricanes than the Atlantic Ocean basin, reflecting the elevated frequency of tropical cyclones in the Pacific region. Figure \ref{fig:hurr_int} (b) and (d) show the hurricane counts by year in the Atlantic Ocean basin, which reveals an upward trend, signifying a consistent increase in hurricanes over the years. This trend line underscores the rising frequency of hurricanes in the Atlantic Ocean, emphasizing the growing impact of these storms on the region. Conversely, there is a less pronounced slope in the trend line in the Pacific Ocean, indicating a relatively stable or slow increase in hurricane counts over the years. While the Pacific basin does experience its share of hurricanes, the trend suggests a more modest increase compared to the Atlantic.

\subsection*{Generation of Natural Language-based Weather Forecast Reports Dataset} \label{natural_language}

% \subsubsection*{Impact}
While there have been many advances in data-driven weather and climate research and forecasting, there are still opportunities in leveraging AI to perform more human-oriented tasks. Further, natural language forecasts are often the primary way that a forecast is communicated to the public. Thus, natural language-based forecasts have a large impact on day-to-day life and play a crucial role in various applications including disasters and severe weather warnings, agricultural planning, and business and economic planning.

For enhanced forecast communication, broadcast meteorologists have been constantly exploring ways for generating interactive forecasts. Generating natural language weather discussions, however, combines computational linguistics with weather analysis, which is a difficult task. The Forecast Generator (FOG) based on the Forecast Production Assistant (FPA) uses rules and natural-langauge generator for converting weather maps into forecast text \citep{294135}. Similarly, GALiWeather \citep{ramos2014linguistic}, SUMTIME-MOUSAM \citep{reiter2005choosing}, and other work \citep{zhang2011meteorological}, implement translating weather data into interactive reports and graphs. However, these frameworks lack standardization and generalizability in the implementation. With recent advances in large-language models (LLMs) and Vision-language models (VLMs), an AI-model can be used to generate forecast outputs after learning the lexical patterns and semantic present in the training data. The proposed novel dataset can be used to train such a model or fine-tune pre-trained FMs to generate textual weather forecast reports based on weather predictions.

Typically, the forecast parameters for generating forecast discussions include daily max and daily min temperatures ($T_m$), relative humidity ($R_h$), wind ($u, v$), cloud cover, probability of precipitation, and weather type. These parameters are used to develop grids of the expected weather conditions. The gridded forecasts are published to the local database and checked for quality. Subsequently, the human forecasters use specialized software to develop products such as the Zone Forecast Prediction, the Area Forecast Matrix, and the Point Forecast Matrix to develop the final textual forecast reports. These forecast reports are typically generated locally, state-wide, and country-wide. 

The proposed dataset is focused on automating this process for generating end-to-end AI-based, natural language weather forecast reports.

\paragraph*{Data Description}
Weather discussion reports involves textual descriptions of the weather, thus, leading to a multi-modal dataset. Table \ref{tab:eda_wx_forecasting} shows certain quantitative stats for the proposed dataset. 

\begin{table}
    \centering
    \caption{Quantitative details of the proposed dataset comprising of subsetted atmospheric states along with the labels. }
    \begin{tabular}{|c|c|c|}
    \hline
        Number of samples & Temporal Span of the forecast labels & Geographical Extent  \\
        \hline
        1249 & August 2014 - December 2017  & CONUS  \\
        \hline
    \end{tabular}
    
    \label{tab:eda_wx_forecasting}
\end{table}

To address the multi-modality present in the dataset due to text labels, we performed following pre-processing steps to the labels (Refer Appendix B).

\begin{itemize}
    \item \textit{Removing punctuation and special characters}: Punctuation characters are important for English grammar. However, they are avoided for text analysis. It is observed that special characters such as the newline character, \textit{\textbackslash n}, are prominently present in the forecast reports. These characters do not enhance information in the context of weather conditions and are removed from the labels.
    \item \textit{Expanding contractions} - Contractions are shortened representations of words such as \textit{don't} for \textit{do not}. The contractions are expanded to understand the full meaning of the text.
    \item \textit{Unifying the word case} - Usually, the forecast reports contain the same words in different casing, such as \textit{THUNDERSTORM} and \textit{thunderstorm}. Since, from a natural language processing perspective, both words represent the same meaning, we unified the word case to lowercase in all the label forecast reports.
    \item \textit{Removing numerals and stop words}: Numerals and stop words (words like \textit{I, we, are, the}) in forecast reports decrease the importance of the domain-specific words (see Figure~\ref{fig:stop_words_a} in Appendix B). Additionally, stop words increase the dataset size and reduce the focus on informative words.  
\end{itemize}

% \subsubsection*{Data Records}
For this task, the High-Resolution Rapid Refresh (HRRR) is used as fine-tuning data along with weather reports as labels. The weather reports are scraped from the Storm Prediction Center's (SPC) archive provided by the National Oceanic and Atmospheric Administration (NOAA). These discussions are national forecasts of severe weather across CONUS.

\begin{figure}
\centering
    \includegraphics[width=0.95\columnwidth]{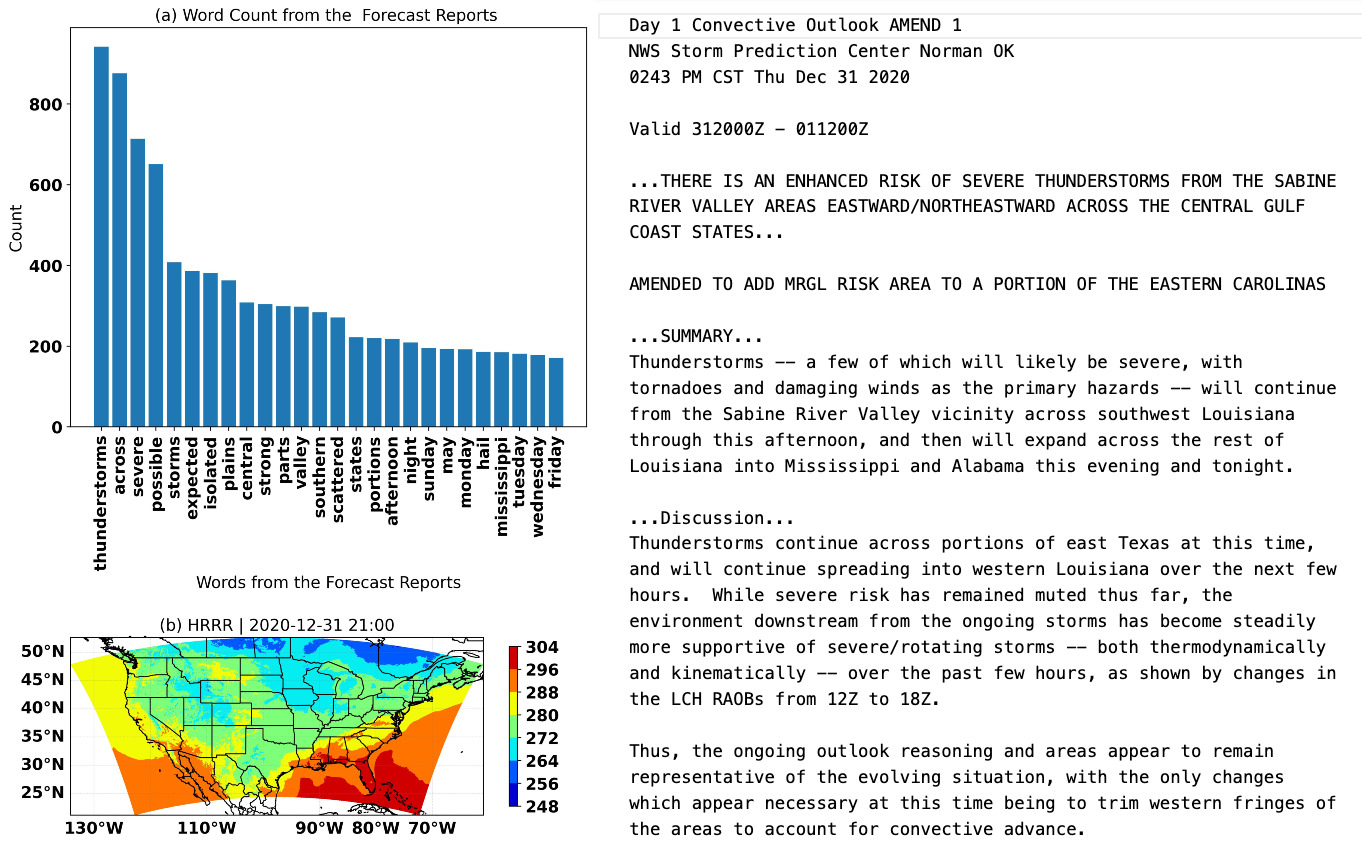}
    \caption{ (a) Illustration of distribution of words in the forecast reports, (b) Illustration of training data for (b) surface temperature and (c) corresponding forecast report for December 31, 2020. Weather data are from the HRRR 21Z analysis on December 31, 2020.}
    \label{fig:wx_report}
\end{figure}

Figure~\ref{fig:wx_report} above illustrates the training data and corresponding forecast report. Each training data point, $x_{i}$, consists of an atmospheric analysis map whereas, weather report published at $0700 hrs$ of the same date as the corresponding label, $y_i$, where $i$ represents an instance of the data sample. Naturally, generating weather forecast discussions is a challenging problem attributed to variability in natural language generation and finding accurate mapping with the corresponding atmospheric phenomenon.

%Technical Validation or Benchmarking comprising of 2 sections for each downstream tasks - Baseline model and results

%Downstream 1

\section*{Technical Validation}
\label{technical_validation}

This section presents the technical validation of the datasets described above for baseline experiments, using both quantitative and qualitative metrics.

\subsection*{Aviation Turbulence Detection} 
\label{sec:AVTvalidation}

The aviation turbulence detection dataset is validated by training an Artificial Neural Network (ANN) to perform a binary classification with two classes - turbulent and not-turbulent, for each of the LOW, MIDDLE, and HIGH levels. The trained ANN has 5 hidden layers of sizes 100, 60, 60, 40, and 20. For activation, Leaky ReLU is used within the hidden layers, and a sigmoid function is used for the classification layer. To mitigate the class imbalance, the loss is weighted by the class proportions. The ANN is evaluated based on the Overall Accuracy (OA), Probability of Detection (PoD) and False Alarm Rate (FAR) as performance metrics. 

\begin{table}[ht]
\centering
\caption{Performance of ANN for aviation turbulence detection based on overall accuracy, probability of detection (PoD), and false alarm rate (FAR)}
\begin{tabular}{lcccc}
\hline
Level   & Accuracy & PoD   & FAR   & F1    \\ \hline
LOW     & 80.0\%   & 56.5\% & 17.8\% & 31.5\% \\
MIDDLE  & 80.4\%   & 38.9\% & 17.3\% & 17.5\% \\
HIGH    & 80.6\%   & 55.6\% & 17.6\% & 28.3\% \\ \hline
\end{tabular}

\label{tab:turbnet_results}
\end{table}

%\subsubsection*{Results}

Table \ref{tab:turbnet_results} presents the quantitative results based on the chosen performance metrics. The high overall accuracy with a lower probability of detection reflects the large imbalances within the training dataset, as reports of turbulence are relatively rare compared to reports of non-turbulence. Due to the highly localized nature of turbulence compared to the relatively coarser grid spacing of MERRA-2, this downstream task is naturally expected to be challenging for DL. 

% Owing to such inherent difficulties present in this downstream task, it could therefore be considered a stress test for DL systems.

%Downstream 2
\subsection*{Gravity Wave Momentum Fluxes}

A proof-of-concept of the capability of DL models to learn and represent the nonlinear subgrid-scale GW evolution presents a new paradigm in their model parameterization development. This possibility is tested and demonstrated here using a baseline ML model. To test the AI-readiness of the data and to gauge the capability of ML models to learn from it, we used the WxC-Bench GW data to train an Attention-Unet \cite{Oktay.etal2018} convolutional neural network.

%\subsubsection*{Results}

%\begin{figure} [H]
%    \centering
%    \includegraphics[scale=0.6]{images_updates/nonlocal_stencil_gw.pdf}
%    \caption{(a) Colors show an example GW packet excited over the Andes and the square grids show three different nonlocal grids over while vectorized input was provided for training three separate ANNs: a single column ANN (black square) which uses information within the one column (1$\times$1) to predict fluxes, a 3$\times$3 grid (green grid) which uses spatially nonlocal information over the immediately neighboring columns to predict the flux in the single column at the center, and a 5$\times$5 grid (brown) which uses information over 25 columns to predict the fluxes in the center column. (b) and (c) Performance of two ANNs, 1$\times$1 and 3$\times$3, trained on one year of GW flux data in the Andean region for two instances. The true resolved flux is shown in black, and the prediction from the 1$\times$1 and 3$\times$3 ANNs is shown in blue and orange, respectively.}
%    \label{fig:gw_results}
%\end{figure}

\begin{figure}[h]
    \centering
    \includegraphics[scale=0.6]{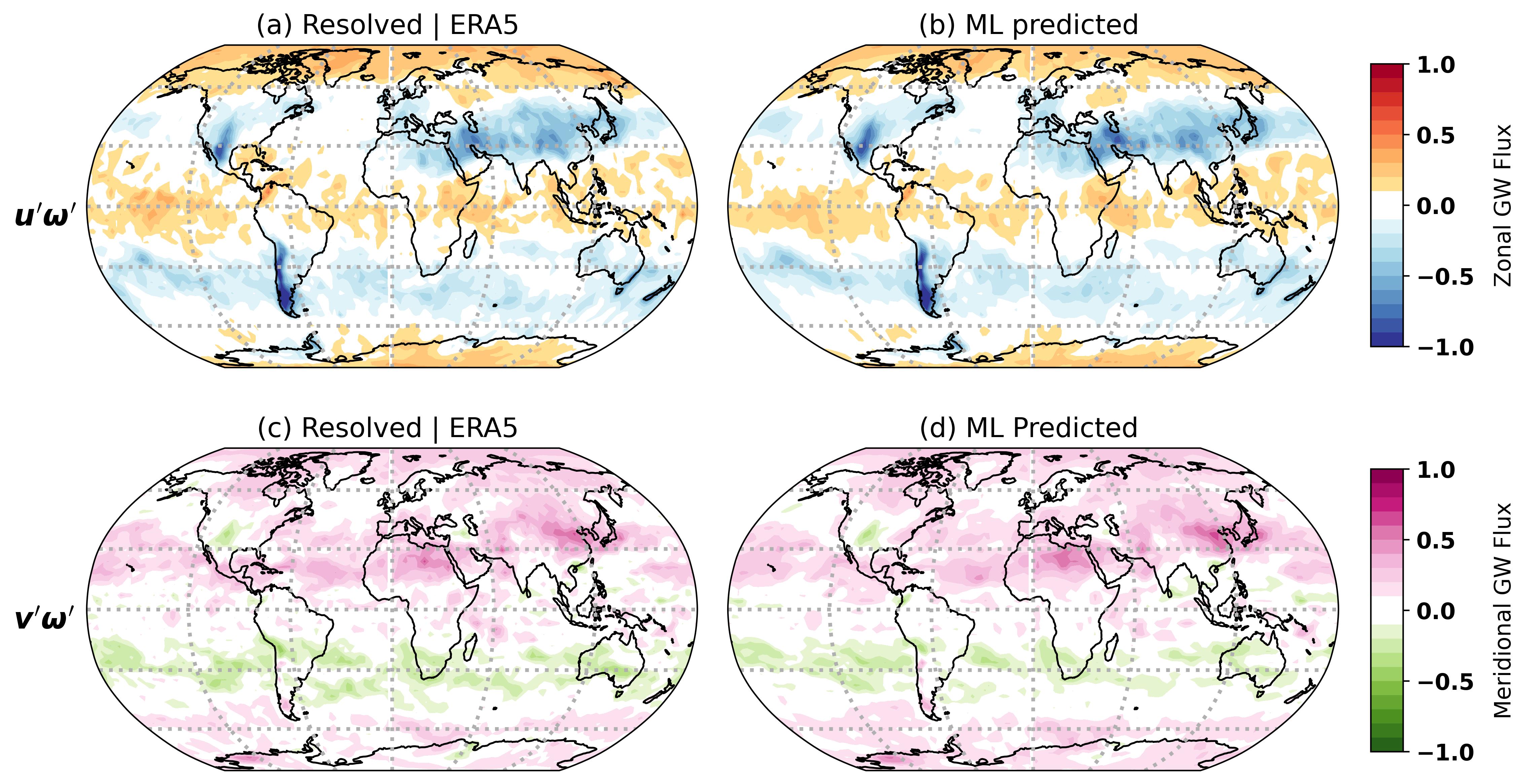}
    \caption{Monthly mean normalized (a) zonal and (c) meridional components of the GW fluxes at 200 hPa (upper troposphere) provided by WxC-Bench for May 2015. (b) and (d) respectively show the predicted zonal and meridional fluxes from a convolutional neural network trained on WxC-Bench data. The CNN skillfully identifies the stationary and non-stationary hotspots and the large- and small-scale features at all latitudes, illustrating the capability of ML models to represent the nonlocal evolution of GWs (and possibly other mesoscale processes in atmosphere models.}
    \label{fig:gw_results}
\end{figure}

Figure \ref{fig:gw_results} shows the CNN's performance in predicting the subgrid-scale fluxes when trained globally on the WxC-Bench GW momentum flux data. The ML model skillfully learns the global fluxes both over the land and the ocean. Over the land, the model learns fluxes over key hotspots including the Rocky mountains, Andes, and East Asian Mountains. Over the ocean, the mean predicted climatology captures the tropical convective fluxes associated with the ITCZ and the midlatitude fluxes from storm tracks over the Southern Ocean. The predictability is notably higher in the midlatitudes and over stationary structures like orography, with $R^2$ values of up to 0.6 in the midlatitude troposphere vs. $R^2$ values of 0.3 in the tropical troposphere (see \cite{Gupta.etal2024b} for a more detailed analysis).

Current operational parameterizations struggle to accurately capture the global distribution of mesoscale momentum fluxes \cite{Geller.etal2013}. In principle, ML models such as this, trained on resolved GW fluxes provided by WxC-Bench, can be subsequently used to represent the learned subgrid-scale fluxes in coarser-climate models that cannot resolve these scales. The flux data can be further used to benchmark or compare various ML-driven GW parameterizations.

%Downstream 3

\subsection*{Weather Analog Search}

\begin{figure}[H]
    \centering
    \includegraphics[width=1\linewidth]{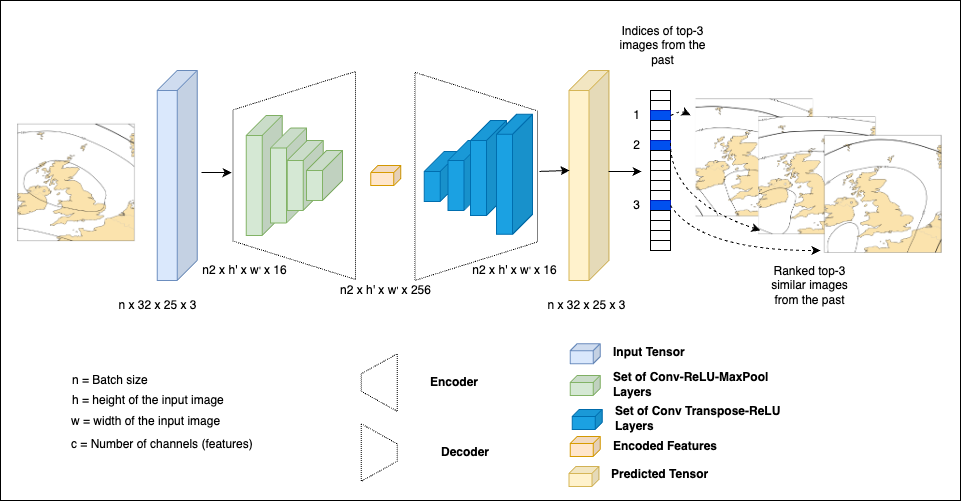}
    \caption{Illustration of the baseline model architecture for deep learning based weather analog search.}
    \label{fig:wx_analog_flow}
\end{figure}

Figure~\ref{fig:wx_analog_flow} shows the architecture for a convolutional encoder-decoder-based baseline model for implementing similarity search based on the MERRA-2 dataset. The model takes weather parameters (such as temperature, sea-level pressure, and wind vectors) as image inputs and searches for images closely representing similar conditions from the past based on cosine similarity. Currently, the technical validation experiment is designed to consider individual weather parameters as images for searching the archive for that particular parameter. In other words, the model can search over either temperature or surface pressure, but not both simultaneously. However, we envision a multi-parameter search for weather conditions by combining search results for individual parameters.

%\subsubsection*{Results}

\begin{figure}[h]
    \centering
    \includegraphics[width=1\linewidth]{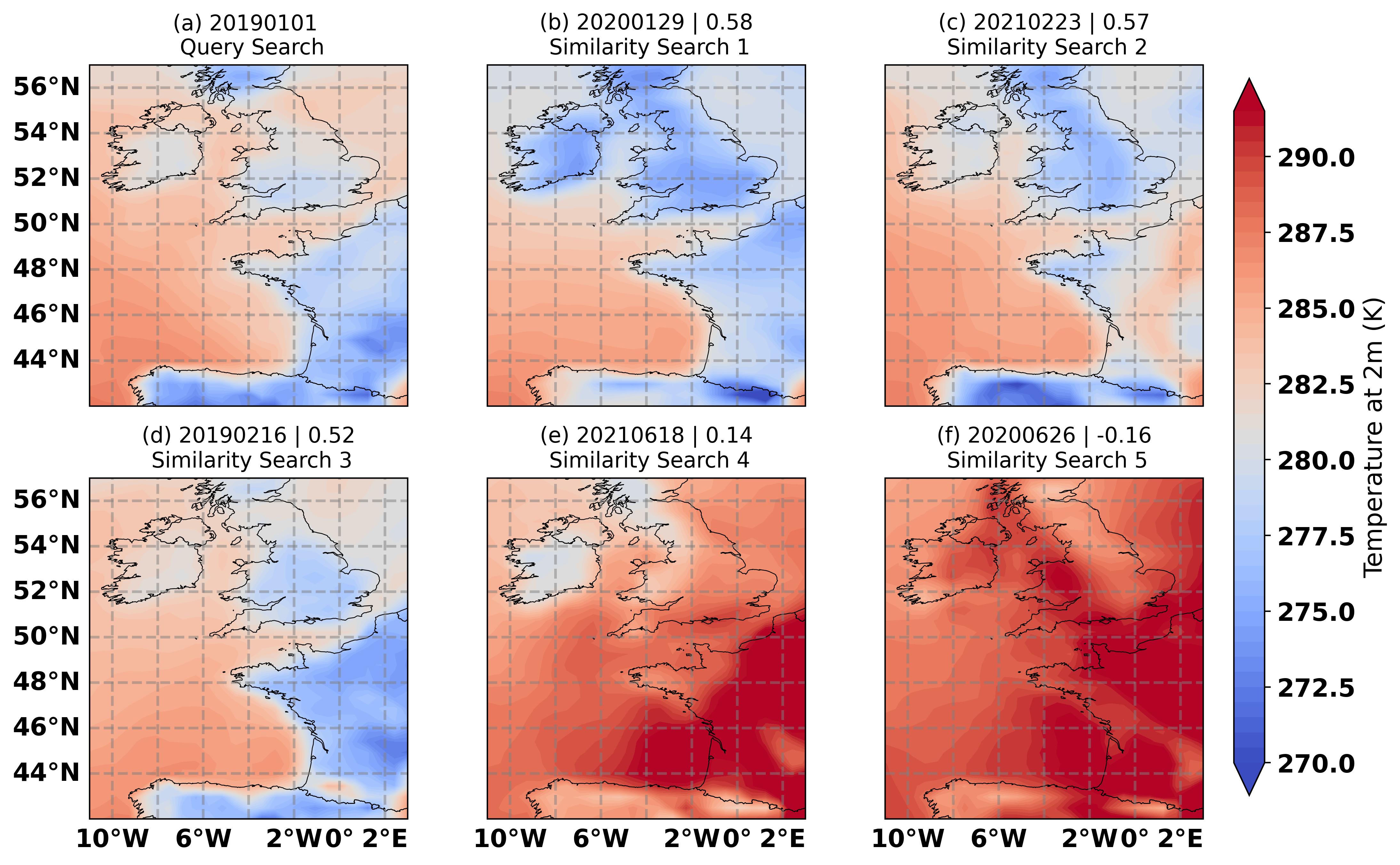}
    \caption{Illustration of the top-5 images for temperature along with their date of occurrence in the past, with their structural similarity score with respect to the query image.}
    \label{fig:wx_analog_results}
\end{figure}

The validation is done by training the encoder-decoder model for 100 epochs on input images of size 32x25. The learning rate is set to 0.001, and the batch size is set to 16 for training. Figure~\ref{fig:wx_analog_results} presents the top five historical images with temperature patterns similar to the query image (Figure 11a), along with their structural similarity scores based on Structural Similarity Index (SSIM)\cite{bouzerdoum2004image}. These SSIM scores range from -1 to 1, with score of 1 indicating an exact similarity, 0 indicating no similarity, and -1 indicating perfect anti-correlation. The first four images retrieved by the similarity search have overall SSIM scores greater than 0.5, while the fourth and fifth images have scores less than 0.2.

%Downstream 4

\subsection*{Long-range Precipitation Forecasting}
\label{sec:s2sValidation}

\begin{figure}
    \centering
    \includegraphics[width=0.5\textwidth]{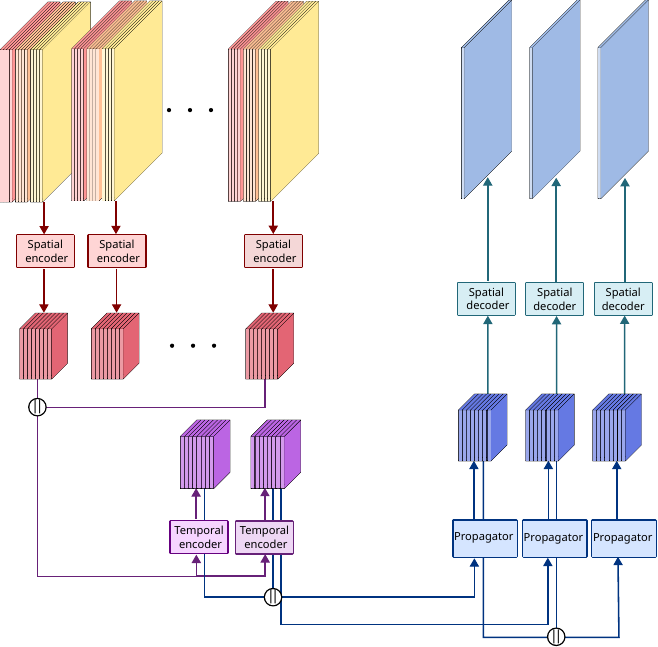}
    \caption{
    The architecture of the convolutional, auto-regressive forecast model used to implement the machine learning baseline model for the long-range precipitation forecast.
    }
    \label{fig:ltp_baseline_model}
\end{figure}

The baseline model for the precipitation forecast task is an auto-regressive forecast model based on a convolutional neural network. The architecture of the model is illustrated in Figure~\ref{fig:ltp_baseline_model}. The model input, which consists of the combined observations from the three satellite datasets from the eight days before the forecast initialization time, is first encoded spatially. The spatial encoder encodes the input from each day independently using separate stems for each observations source. Each stem consists of a $3 \times 3$ convolution block with 64 output channels. The resulting stem-outputs from the three observation sources are then concatenated and fed into the shared spatial encoder, which encodes the combined observations from each day spatially, successively reducing the resolution by a factor of 4 in two stages. The spatially encoded observations are then concatenated and passed through two separate temporal encoders that map the encoded observations into the two initial latent forecast states. From these two states, successive latent forecast states are produced using a propagator module consisting of an encoder-decoder architecture that takes the states from two previous steps and maps them to a latent state representing the forecast for the following day. The predicted states are decoded using a shared decoder, which upsamples the forecasts back to the original resolution. Finally, a head consisting of a single point-wise convolution layer is used to produce precipitation forecast for all predicted forecast steps.

All components of the model, except the stems and head, consist of ResNeXt \citep{xie_17_resnext} blocks. The exact configuration of components is provided in Table~\ref{tab:ltp-baseline_config}. Inputs to each convolution are padded to conserve the image size. Since the observations are global, periodic padding is applied along the longitudinal dimension, while reflect padding is applied along the latitudinal dimension.

The baseline is trained to predict 32 quantiles of the posterior distribution using a quantile loss function \citep{pfreundschuh2018neural}. The training uses the AdamW optimizer \citep{loshchilov_17} with an initial learning rate of $10^{-4}$ and a cosine annealing learning-rate schedule over 50 epochs. The model was trained to predict precipitation up to 16 days into the future. To make the model robust to changes in the input observations, individual input channels are dropped with a probability of $2\unit{\%}$ and all observations from a sensor are dropped with a probability of $5\unit{\%}$.

%\subsubsection*{Results}

\begin{figure}
    \centering
    \includegraphics[width=1\textwidth]{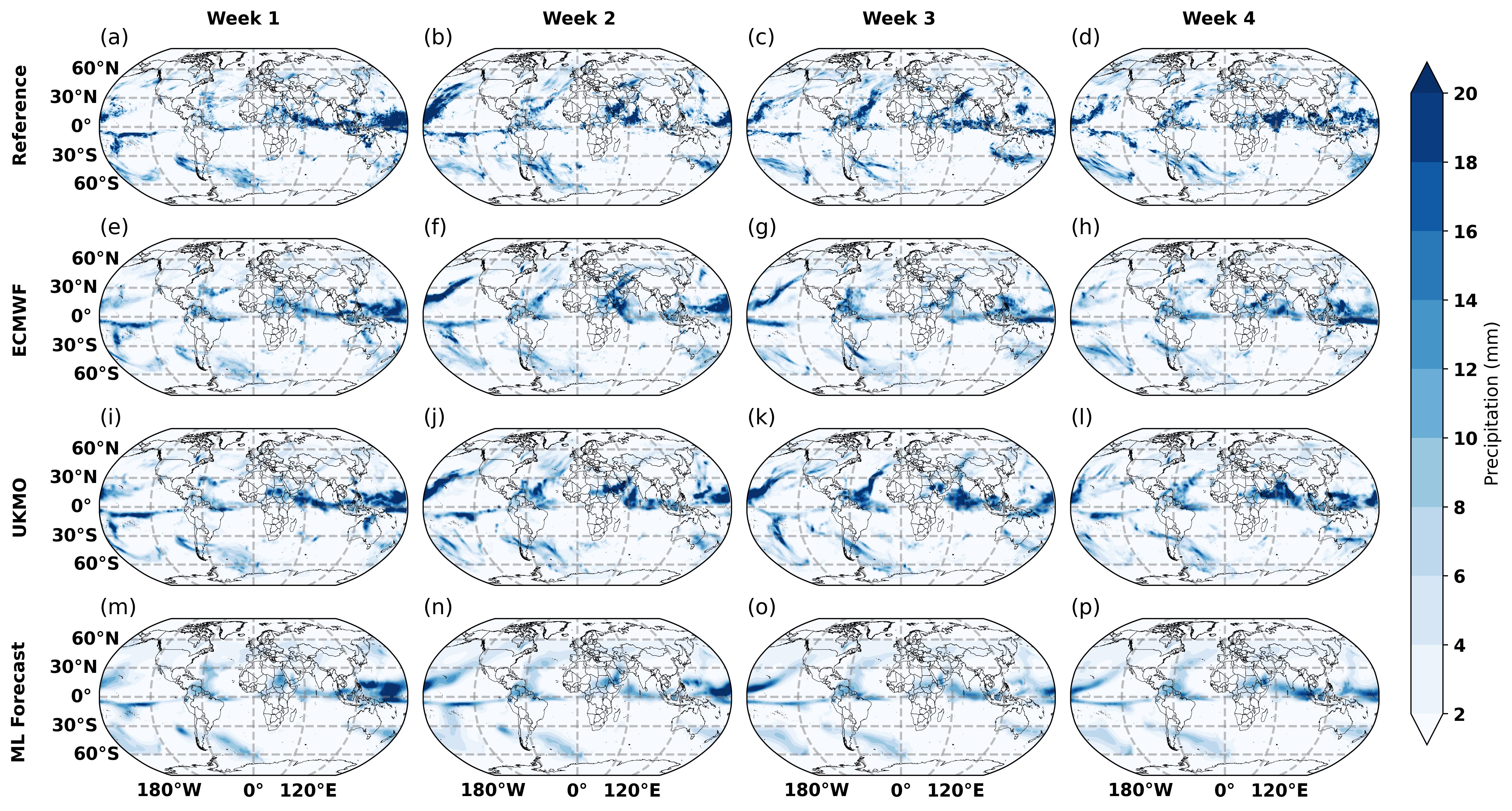}
    \caption{Reference and predicted mean daily precipitation for forecasts initialized on 10 January 2019. The first row shows the satellite-based reference precipitation estimates from the IMERG Final product. Rows two and three show the corresponding forecasts from the control members of the ECWMF and UKMO forecasts from the S2S database, respectively. The last row shows the forecasts from the machine learning-based baseline model.}
    \label{fig:ltpf_example_predictions}
\end{figure}

The ML forecasts are compared to conventional NWP reference forecasts obtained from the S2S-database \citep{vitart_17_s2s_database}, which archives forecasts from various operational centers. For this evaluation, we consider only the forecasts from the ECMWF and the United Kingdom’s Met Office (UKMO) as these were found to be among the best-performing models in an evaluation of surface precipitation forecasts \citep{andrade_19_s2s_evaluation}. Examples of forecasts initialized on 10 January 2019 are shown in Figure~\ref{fig:ltpf_example_predictions}. The ECMWF and UKMO forecasts yield good agreement with the reference estimates during the first week but yield a progressively larger error for the following weeks. Our machine learning-based baseline forecast also successfully captures the principal precipitation features during the first weeks but its predictions are generally less well-defined than those of the conventional models.

\begin{figure}
    \centering
    \includegraphics[width=\textwidth]{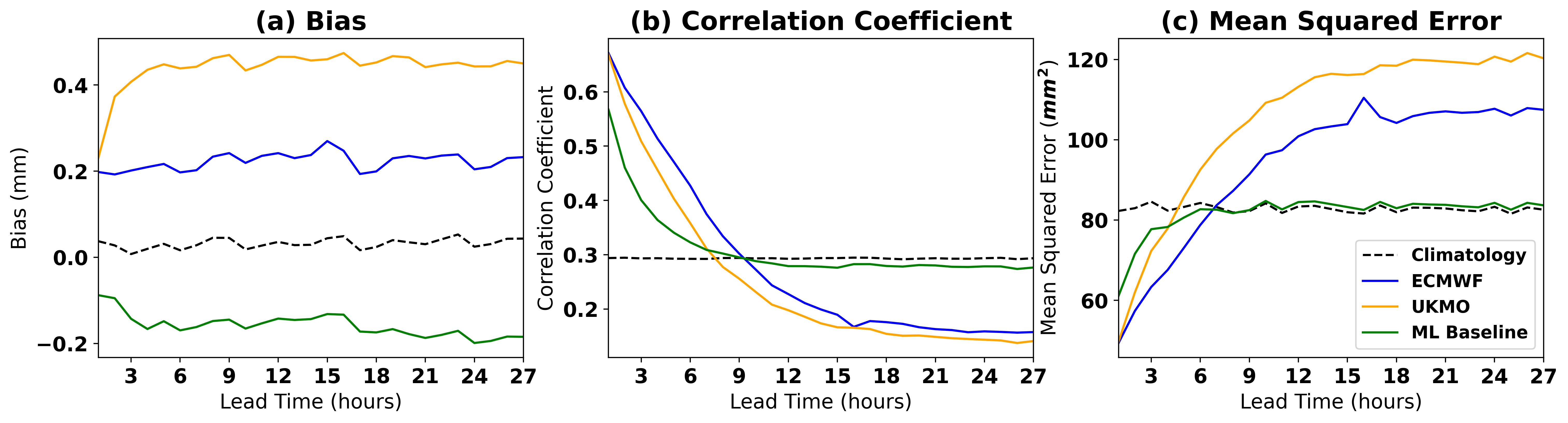}
    \caption{
    Accuracy of daily accumulated precipitation forecasts. Panel (a) shows the forecast bias defined as the area-weighted mean of the difference between predicted and reference precipitation. Panel (b) shows the area-weighted linear correlation coefficient. Panel (c) shows the area-weighted mean-squared error.
    }
    \label{fig:ltpf_baseline_metrics}
\end{figure}

The results of a numerical evaluation of all forecasts from the evaluation period 1 January 2019 and 31 December 2022 are displayed in Figure~\ref{fig:ltpf_baseline_metrics}. The forecasts are evaluated using the area-weighted forecast bias, linear correlation coefficient, and mean-squared error with respect to the reference precipitation estimates. In terms of bias, the ML-based model has a relatively small dry bias while the conventional models exhibit wet biases. The ML baseline forecasts is closest to the reference data, which is expected given that the precipitation estimates it was trained on were derived from the same precipitation product that is used for the evaluation. In terms of correlation coefficient, the conventional models yield higher correlations at shorter times but their advantage over the ML model decreases with longer lead times. The correlation coefficient of the conventional forecasts falls below that of the ML baseline after lead times exceeding 10 days but at this time the accuracy has decreased to or below that of a monthly precipitation climatology. Similar tendencies can be observed in the results of the mean-squared error.

%Downstream 5

\subsection*{Hurricane Forecasting based on FourCastNet} 
\label{sec:HURvalidation}

For validating the hurricane forecasting dataset, we gathered and analyzed a total of 1,459 named hurricanes that formed within the time frame of 1980 to 2020. This dataset includes 649 hurricanes from the Atlantic Ocean basin and 810 hurricanes from the Pacific Ocean basin. 

\begin{figure}[H]
\noindent\includegraphics[width=\textwidth]{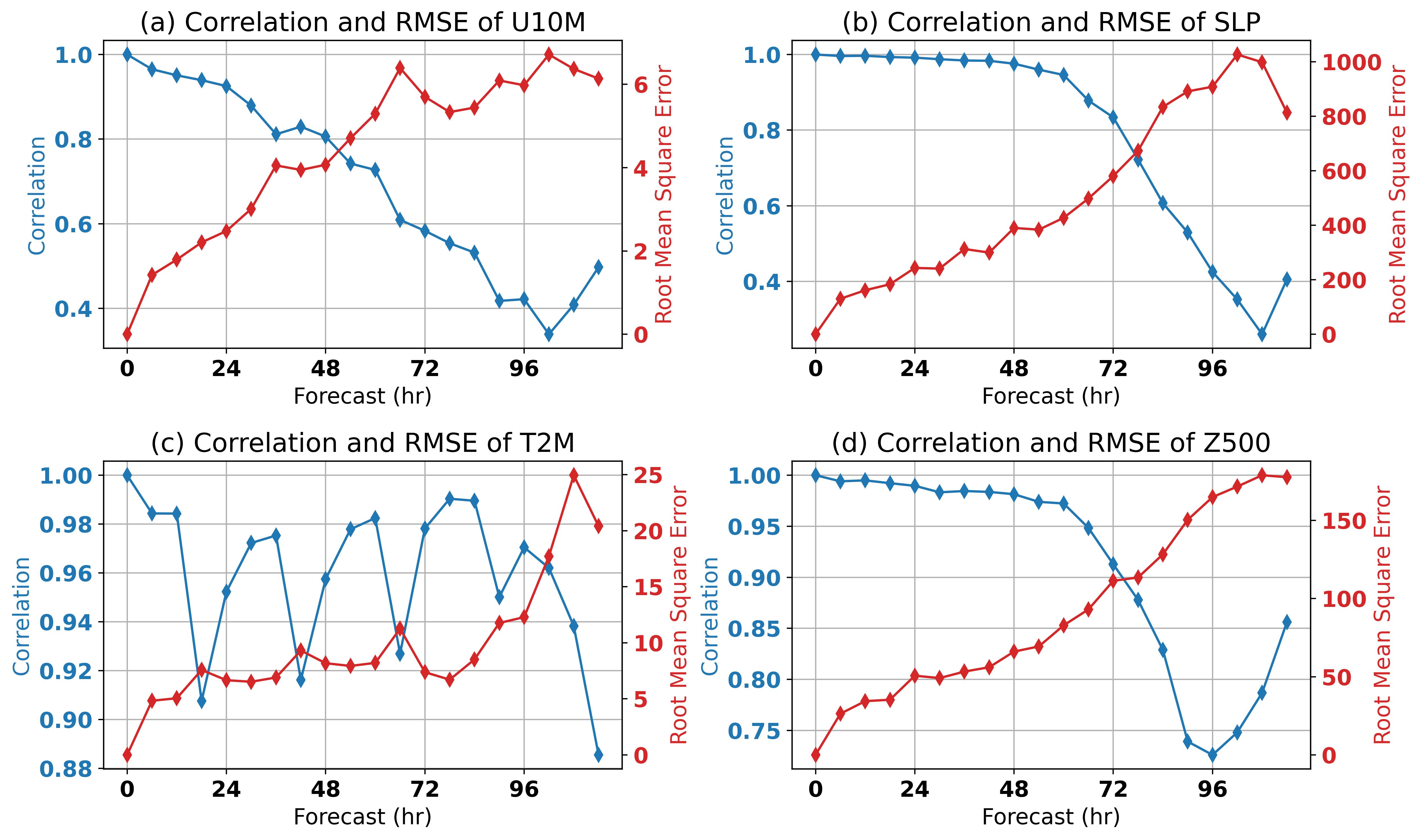}
\caption{Time series of the correlation and root mean square error of wind speed (m/s), SLP (Pa), T2M (K) and Z500 (m) for the hurricane Michael (2018) over the hurricane Michael domain (105$^{\circ}$W, 60$^{\circ}$W, 20$^{\circ}$N, 50$^{\circ}$N ).}
\label{hurricane_baseline01}
\end{figure}

To validate this task, FourCastNet has been used as the baseline model, which is a neural network architecture specifically designed for high-resolution forecasting tasks by combining elements from the Fourier Neural Operator (FNO) learning approach and the ViT architecture \cite{kurth2023fourcastnet}. The model effectively addresses the computational challenges associated with high-resolution data by framing spatial token mixing as a continuous global convolution, which is efficiently implemented in the Fourier domain using Fast Fourier Transforms (FFTs). This innovative approach significantly reduces spatial mixing complexity, making FourCastNet well-suited for high-resolution forecasting. 

%\subsubsection*{Results}

Figure \ref{hurricane_baseline01} shows an evaluation of the model's forecasting performance for key meteorological parameters over 5 days for Hurricane Michael. Wind speed (WS) exhibits strong correlations, with values of 0.8 at a 2-day forecast, 0.7 at a 3-day forecast, and 0.4 at a 5-day forecast, indicating a significant decline in accuracy over the extended period. Sea level pressure (SLP) shows high correlations, particularly at shorter forecast intervals, with a remarkable 0.97 at a 2-day forecast, decreasing to 0.82 at 3 days, and further reducing to 0.3 at the 5-day mark. The temperature at 2 m above ground level (T2M) displays diurnal variation in correlation, with daytime values peaking around 0.98 and nighttime values maintaining around 0.90. T2M outperforms other variables, consistently showcasing superior performance. Geopotential height at 500 hPa (Z500) is a reliable predictor, with correlations of 0.97 at 2 days, 0.92 at 3 days, and 0.75 at 4 days. The decreasing correlation values over the forecast period suggest a slow reduction in accuracy, highlighting the importance of considering shorter time frames for more reliable predictions. These findings collectively underscore the variable-dependent nature of the model's accuracy and emphasize the significance of temporal considerations. The time series correlation and root mean-squared error (RMSE) analysis show the nuanced forecasting capabilities of the model across different variables and forecast durations. 

\begin{figure}[t]
\noindent\includegraphics[width=\textwidth]{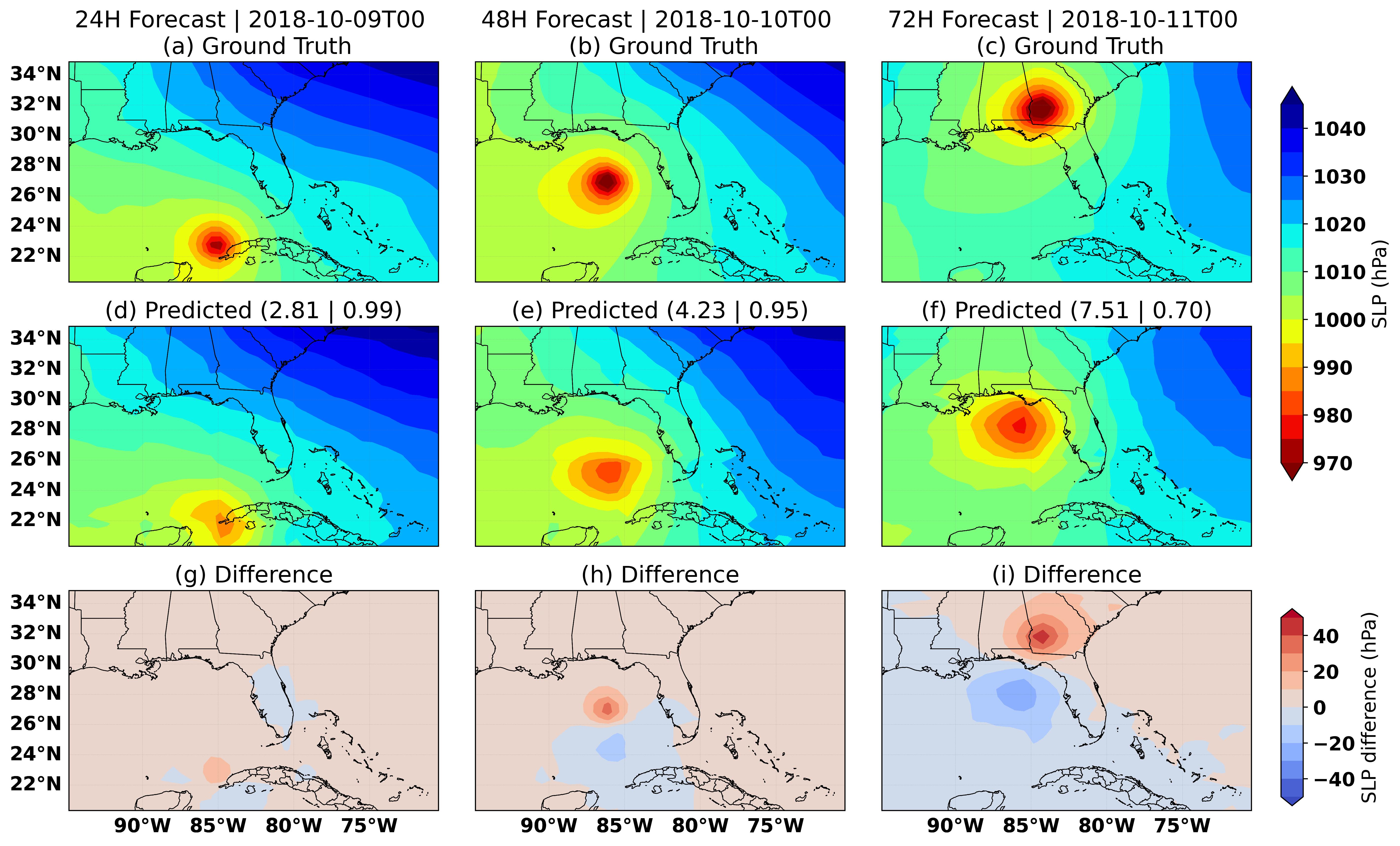}
\caption{Spatial distribution of sea level pressure (Pa) for Hurricane Michael (2018) for 24H, 48H and 72H forecast from the ground truth, prediction, and the difference between the prediction and the ground truth.
}
\label{hurricane_baseline02}
\end{figure}

The spatial distribution of model performance regarding the prediction of Hurricane Michael's intensity in terms of mean sea level pressure (MSLP) for 1-day, 2-day, and 5-day forecasts, as shown in Figure \ref{hurricane_baseline02} from the MERRA trained model. The RMSE for the 1-day forecast was 118.15 Pa, indicating a close alignment of the predicted hurricane center with the ground truth. However, for the 2-day and 3-day forecasts, the RMSE increased to 189.35 and 281.86 Pa, accompanied by correlation coefficients of 0.98 and 0.83, respectively.

Despite the spatial mean RMSE increasing to 281 for the 3-day forecast, it reveals that this error is most pronounced at the hurricane's center. This central discrepancy holds significant implications for hurricane prediction accuracy. The MSLP error for the 2-day and 3-day forecasts is quantified at 35 hPa and 52 hPa, respectively. While the tracking error remains minimal (below 40-50 km), the intensity error is notable, deviating by approximately 50-60 hPa.

\subsection*{Generation of Natural Language-based Weather Forecast Reports}

\begin{figure}[H]
    \centering
    \includegraphics[width=1\linewidth]{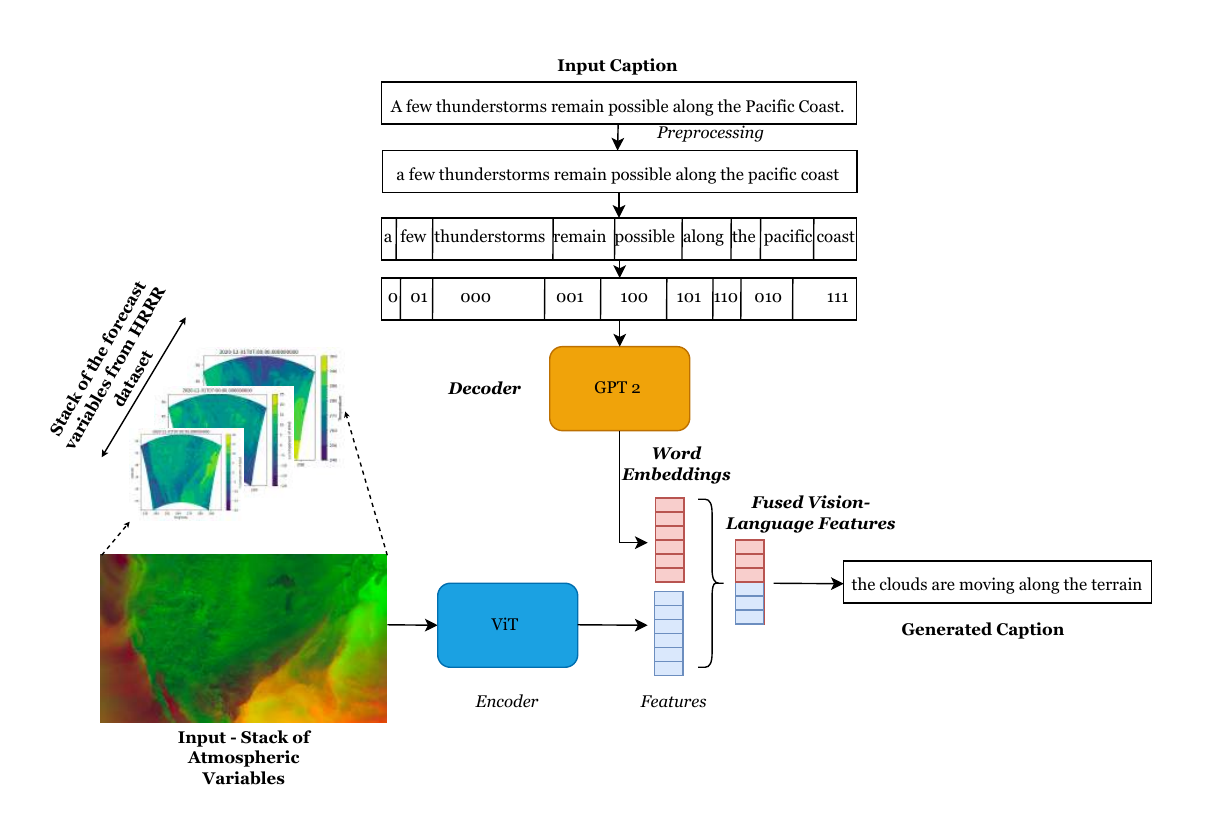}
    \caption{Illustration of high-level architecture for generating natural language-based weather forecast captions. The captions are generated from an encoder-decoder architecture, pre-trained in vision and language models, and fine-tuned to the proposed dataset.}
    \label{fig:wx_forecasting_arch}
\end{figure}

Figure~\ref{fig:wx_forecasting_arch} shows the high-level architecture for technical validation of the proposed dataset. The baseline experiment is performed by fine-tuning an encoder-decoder model for vision-language modeling. The encoder is based on the vision transformer (ViT) \citep{dosovitskiy2020image}, whereas GPT-2 \citep{radford2019language} is used as a decoder model for generating natural language-based forecast reports. The idea is to validate the proposed dataset as an ML-ready dataset capable of training on a multi-modal vision-language-based captioning problem. For the baseline model training, we have extracted the summary from the original forecast reports as labels for training instead of using the entire report. This summary is present in every forecast report under the section highlighted by "...SUMMARY...".

%\subsubsection*{Results}

\begin{table}[H]
    \caption{Quantitative scores for the weather forecast report generation benchmarking.}
    \centering
    \begin{tabular}{ *{3}{c} }
      Date of the weather condition & ROUGE-L score\\ \hline
        2017-03-30  & 0.30  \\
        2018-08-03  & 0.28  \\
        2019-08-30  & 0.42  \\
    \end{tabular}
    \label{tab:bleu_rouge_score}
\end{table}

We validate the model quantitatively based on the Recall-Oriented Understudy for Gisting Evaluation-Longest Common Subsequence (ROUGE-L) score. More details about the training are present in the Appendix A. Table~\ref{tab:bleu_rouge_score} highlights the quantitative evaluation of the proposed architecture based on the selected metrics. We observe a maximum ROUGE-L score of 0.42 across all the test samples. Typically, a ROUGE-L score of 0.4-0.6 is considered decent for image captioning task \citep{lin-2004-rouge}. However, as a future work we are investigating comprehensive evaluation of the generated reports.

\begin{figure}[h]
    \centering
    \includegraphics[width=1\linewidth]{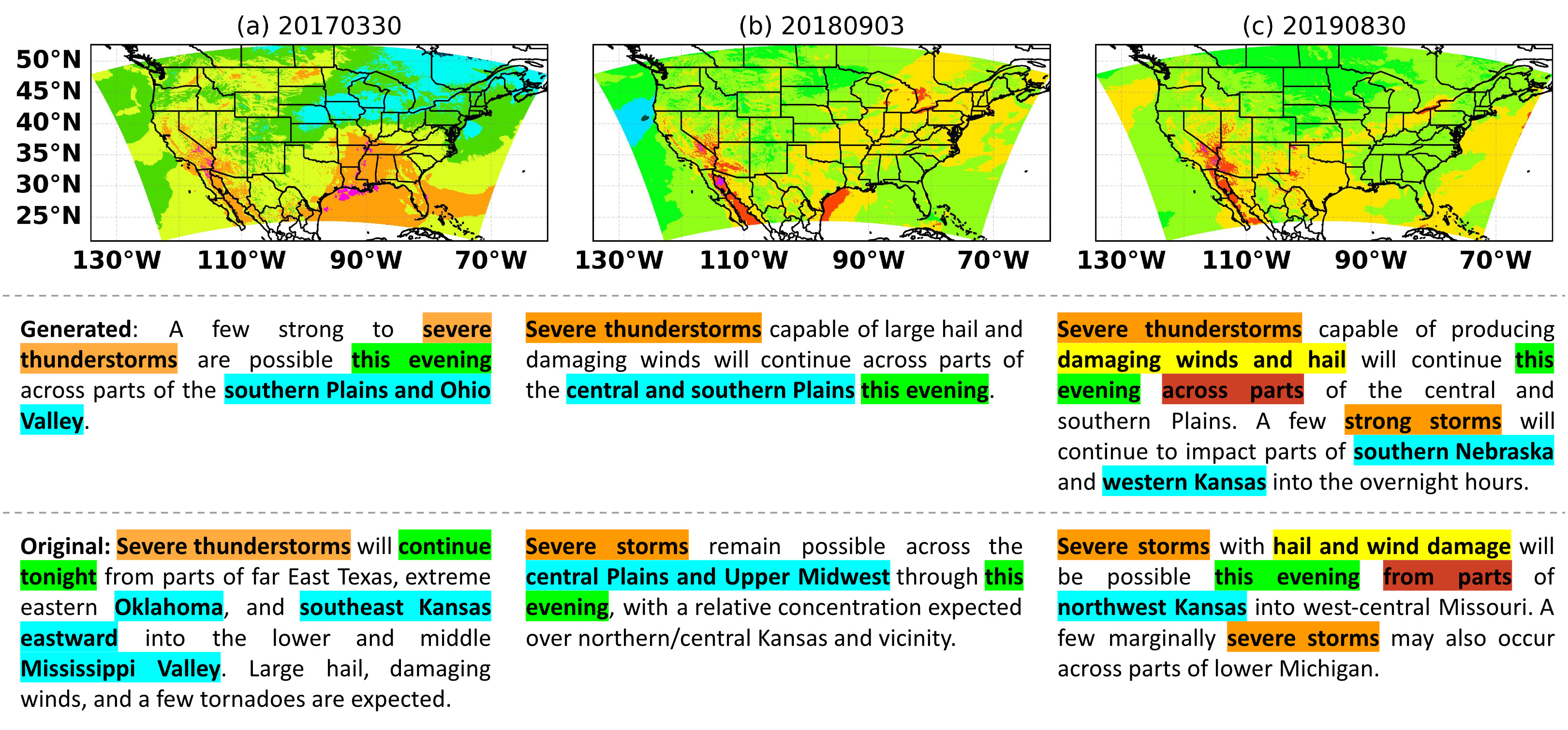}
    \caption{Results show the performance of the natural language-based weather forecasting trained on the proposed dataset. The top row shows an illustration of three channels stacked from temperature, \textit{u}, and \textit{v} components of the wind; the second row shows the generated caption, whereas the third row shows the original caption.}
    \label{fig:forecast-result}
\end{figure}

% \begin{figure}
%     \centering        
%     \includegraphics[width=1\linewidth]{images_updates/wx-forecasting-results.png}
%     \caption{Results show the performance of the natural language-based weather forecasting trained on the proposed dataset. The top row shows an illustration of three channels stacked from temperature, \textit{u}, and \textit{v} components of the wind; the second row shows the generated caption, whereas the third row shows the original caption.}
%     \label{fig:wx-forecast-output}
% \end{figure}

\definecolor{lightblue}{RGB}{0,255,255}
\definecolor{redberry}{RGB}{205,64,37}
\definecolor{lightgreen}{RGB}{0,255,0}
\definecolor{lightorange}{RGB}{255,153,0}

Figure~\ref{fig:forecast-result} shows the generated weather forecast reports along with the original reports. As evident from the results, the baseline model is capturing crucial information such as extreme events information (\colorbox{lightorange}{severe thunderstorms}), geographical details (\colorbox{lightblue}{southern Plains and Ohio Valley}), temporal details (\colorbox{lightgreen}{this evening}), catastrophic event details (\colorbox{yellow}{damaging winds and hail}), and relational details (\colorbox{redberry}{across parts}). However, comprehensive evaluation of the generated captions is being explored as a future work. 

\section*{Conclusion}

Given the tremendous advancements unlocked by the Artificial Intelligence (AI) based Foundation Models (FMs) and Large Language Models (LLMs) in Natural Language Processing and Computer Vision, it seems reasonable to expect a similar impact within the earth and atmospheric sciences. In this context, Machine Learning-ready datasets serve as a foundation for developing new models or fine-tuning existing models for downstream tasks. However, it is crucial that these datasets sample a wide range of realistic and relevant applications as well as data modalities to challenge AI model developers and demonstrate model capabilities to domain-specific scientists.
The primary contribution of this paper - WxC-Bench, is a collection of high-quality datasets corresponding to a broad range of weather- and climate-related downstream tasks. WxC-Bench consists of datasets curated from satellite observations, reanalysis, high-resolution forecasting models, hurricane databases, and even pilot reports. Thus, the dataset can be used for benchmarking and testing the generalization of other specialized AI-models, and assess their ability to handle multi-modal data.

The selected six downstream tasks related to weather and atmospheric sciences are formulated as machine learning problems - classification, prediction, vector space search, and multi-modal language generation. The proposed WxC-Bench, thus, contains pre-processed ML-ready data for a range of atmospheric processes ranging from local scale (aviation turbulence) to synoptic scales (natural language forecasting). WxC-Bench is designed to be used for developing future weather and climate foundation models capable of working with multi-modal data, which can be used not only by researchers, but also by the broader scientific community. The openly-available data repository also promotes future additions of more downstream tasks to augment the capabilities of existing and new AI models trained on WxC-Bench.

% \subsubsection*{Usage Notes}
\section*{Code availability}
\label{code}
Source code to create the datasets - \url{https://github.com/NASA-IMPACT/WxC-Bench}

Dataset Availability - Link to access the dataset on Hugging Face - \url{https://huggingface.co/datasets/nasa-impact/WxC-Bench}

%==========================================================================%

\bibliography{sample, references}

\section*{Acknowledgements}

% Acknowledgements should be brief, and should not include thanks to anonymous referees and editors, or effusive comments. Grant or contribution numbers may be acknowledged.

This work was supported by NASA’s Office of Chief Science Data Officer and Earth Science Division’s Earth Science Scientific Computing, Earth Science Data Systems Program, and the Earth Science Modeling and Analysis Program.

The long-term precipitation forecasting task uses \href{https://confluence.ecmwf.int/display/S2S}{S2S}. S2S is a joint initiative of the World Weather Research Programme (WWRP) and the World Climate Research Programme (WCRP). The original S2S database is hosted at ECMWF as an extension of the TIGGE database. AG and AS were supported by Schmidt Sciences, LLC, a philanthropic initiative founded by Eric and Wendy Schmidt, as part of the Virtual Earth System Research Institute (VESRI). AS acknowledges support from the National Science Foundation through grant OAC-2004492. 

\section*{Competing interests}
The author declares no competing interests.

\section*{Appendix A}
\label{appendix:a}
\subsubsection*{Accessing the dataset}

The proposed WxC-Bench dataset is hosted on the Hugging Face datasets as described in the Code availability \ref{code} section. The dataset can be accessed programmatically using the Hugging Face \lstinline{datasets} library as shown below. 

For all the downstream taks except the natural language based weather report generation, the \lstinline{datasets} variable will have paths to the NetCDF or HDF5 files which can be read using Python libraries such as \lstinline{xarray} for retrieving the data values. For the natural language based weather report generation task, the textual captions can be read using \lstinline{pandas} library, while the grib2 files can be read using Python libraries such as \lstinline{Pygrib}. 

\subsubsection*{Installing the Hugging Face datasets Library}

\begin{mdframed}[backgroundcolor=black!5,leftmargin=0.5cm,skipabove=0.3cm,hidealllines=true,%
  innerleftmargin=0.1cm,innerrightmargin=0.2cm,innertopmargin=-0.0cm,innerbottommargin=-0.10cm]
\begin{lstlisting}[language=Python]
pip install datasets
\end{lstlisting}
\end{mdframed}

\subsubsection*{Cloning the WxC-Bench Repository}

\begin{mdframed}[backgroundcolor=black!5,leftmargin=0.5cm,skipabove=0.3cm,hidealllines=true,%
  innerleftmargin=0.1cm,innerrightmargin=0.2cm,innertopmargin=-0.0cm,innerbottommargin=-0.10cm]
\begin{lstlisting}[language=Python]
git clone https://huggingface.co/datasets/nasa-impact/WxC-Bench WxC-Bench
cd WxC-Bench
\end{lstlisting}
\end{mdframed}

\subsubsection*{Loading the Dataset for Aviation Turbulence Detection Task}

\begin{mdframed}[backgroundcolor=black!5,leftmargin=0.5cm,skipabove=0.3cm,hidealllines=true,%
  innerleftmargin=0.1cm,innerrightmargin=0.2cm,innertopmargin=-0.0cm,innerbottommargin=-0.10cm]
\begin{lstlisting}[language=Python]
from datasets import load_dataset

datasets = load_dataset("./WxC-Bench/aviation_turbulence/dataset.py", trust_remote_code=True)
\end{lstlisting}
\end{mdframed}

\subsubsection*{Loading the Dataset for Natural Language based Weather Report Generation Task}

\begin{mdframed}[backgroundcolor=black!5,leftmargin=0.5cm,skipabove=0.3cm,hidealllines=true,%
  innerleftmargin=0.1cm,innerrightmargin=0.2cm,innertopmargin=-0.0cm,innerbottommargin=-0.10cm]
\begin{lstlisting}[language=Python]
from datasets import load_dataset

datasets = load_dataset("./WxC-Bench/weather_forecast_discussion/dataset.py", trust_remote_code=True)
\end{lstlisting}
\end{mdframed}

\subsubsection*{Loading the Dataset for Gravity Wave Parameterization Task}

\begin{mdframed}[backgroundcolor=black!5,leftmargin=0.5cm,skipabove=0.3cm,hidealllines=true,%
  innerleftmargin=0.1cm,innerrightmargin=0.2cm,innertopmargin=-0.0cm,innerbottommargin=-0.10cm]
\begin{lstlisting}[language=Python]
from datasets import load_dataset

datasets = load_dataset("./WxC-Bench/nonlocal_parameterization/dataset.py", trust_remote_code=True)
\end{lstlisting}
\end{mdframed}

\subsubsection*{Loading the Dataset for Long-term Precipitation Forecasting Task}

\begin{mdframed}[backgroundcolor=black!5,leftmargin=0.5cm,skipabove=0.3cm,hidealllines=true,%
  innerleftmargin=0.1cm,innerrightmargin=0.2cm,innertopmargin=-0.0cm,innerbottommargin=-0.10cm]
\begin{lstlisting}[language=Python]
from datasets import load_dataset

datasets = load_dataset("./WxC-Bench/long_term_precipitation_forecast/dataset.py", trust_remote_code=True)
\end{lstlisting}
\end{mdframed}

\subsubsection*{Loading the Dataset for Hurricane Detection Task}

\begin{mdframed}[backgroundcolor=black!5,leftmargin=0.5cm,skipabove=0.3cm,hidealllines=true,%
  innerleftmargin=0.1cm,innerrightmargin=0.2cm,innertopmargin=-0.0cm,innerbottommargin=-0.10cm]
\begin{lstlisting}[language=Python]
from datasets import load_dataset

datasets = load_dataset("./WxC-Bench/hurricane/dataset.py", trust_remote_code=True)
\end{lstlisting}
\end{mdframed}

\section*{Appendix B}
\label{appendix:b}

\subsubsection*{Architecture for ML long-term precipitation forecasts}
\begin{table}[h]
    \caption{
    Model configuration for the machine learning baseline of the long-range precipitation forecast. The table lists the stages for each component of the model. Each stage comprises one or multiple blocks, and the table lists their respective configuration and number of repetitions. $k$ denotes the kernel size of the convolution operation, while $b$ is the bottleneck width, and $c$ the cardinality following \citep{xie_17_resnext}.
    }
    \begin{tabular}{|cccccc|}
    \hline 
    \multicolumn{6}{|c|}{Spatial encoder} \\
    \hline 
    Stage & Input channels & Output channels & Up/Downsampling & Block type & \# blocks \\
    Head & 24 / 48 / 14 & 64 & - & Basic convolution, $k = 3$ & 1 \\
    Aggregator & 192 & 128 & - & Basic convolution, $k =1$, $k = 1$   & 1\\
    Stage 1 & 128 & 192  & - & ResNeXt, $k = 3$, $b = 3$, $c = 32$ & 2 \\
    Stage 2 & 192 & 256 & down, 2 & ResNeXt, $k = 3$, $b = 4$, $c = 32$  & 2 \\
    Stage 3 & 256 & 256 & down, 2 & ResNeXt, $k = 3$, $b = 4$, $c = 32$  & 2 \\
    \hline 
    \multicolumn{6}{|c|}{Temporal encoder} \\
    \hline
    Stage 1 & 2048 & 1024 & - & ResNeXt, $k = 3$, $b = 32$, $c = 32$ & 1 \\
    Stage 2 & 1024 & 512 & - & ResNeXt, $k = 3$, $b = 16$, $c = 32$  & 1 \\
    \hline 
    \multicolumn{6}{|c|}{Propagator} \\
    \hline
    Stage 1 & 512 & 1024 & - & ResNeXt, $k = 3$, $b = 32$, $c = 32$ & 3 \\
    Stage 2 & 1024 & 1024 & down, 2 & ResNeXt, $k = 3$, $b = 32$, $c = 32$  & 3 \\
    Stage 3 & 1024 & 1024 & down, 2 & ResNeXt, $k = 3$, $b = 32$, $c = 32$  & 3 \\
    Stage 4 & 1024 & 1024 & up, 2 & ResNeXt, $k = 3$, $b = 32$, $c = 32$  & 3 \\
    Stage 5 & 1024 & 512 & up, 2 & ResNeXt, $k =3$,$b = 16$, $c = 32$  & 3 \\
    \hline 
    \multicolumn{6}{|c|}{Decoder} \\
    \hline
    Stage 1 & 512 & 256 & up, 2 & ResNeXt, $k = 3$, $b = 8$, $c = 32$ & 3 \\
    Stage 2 & 256 & 128 & up, 2 & ResNeXt, $k = 3$, $b = 4$, $c = 32$  & 3 \\
    Head & 128 & 32 & - & Basic convolution, $k = 1$ & 1 \\
    \hline 

    \end{tabular}
    \label{tab:ltp-baseline_config}
\end{table}

\section*{Appendix C}
\label{appendix:c}
\subsubsection*{Exploratory Data Analysis for the Natural Language Generation Downstream Task}

The exploratory data analysis (EDA) presents statistical and empirical insights about the data, helping to understand the implicit probabilistic distribution and biases. For this task, we comprehensively analyze the forecast reports and present EDA for all the text from reports accumulated over a year (2019-20).

\begin{figure}[ht]
    \centering
    \begin{subfigure}
        \centering
        \includegraphics[width=1\linewidth]{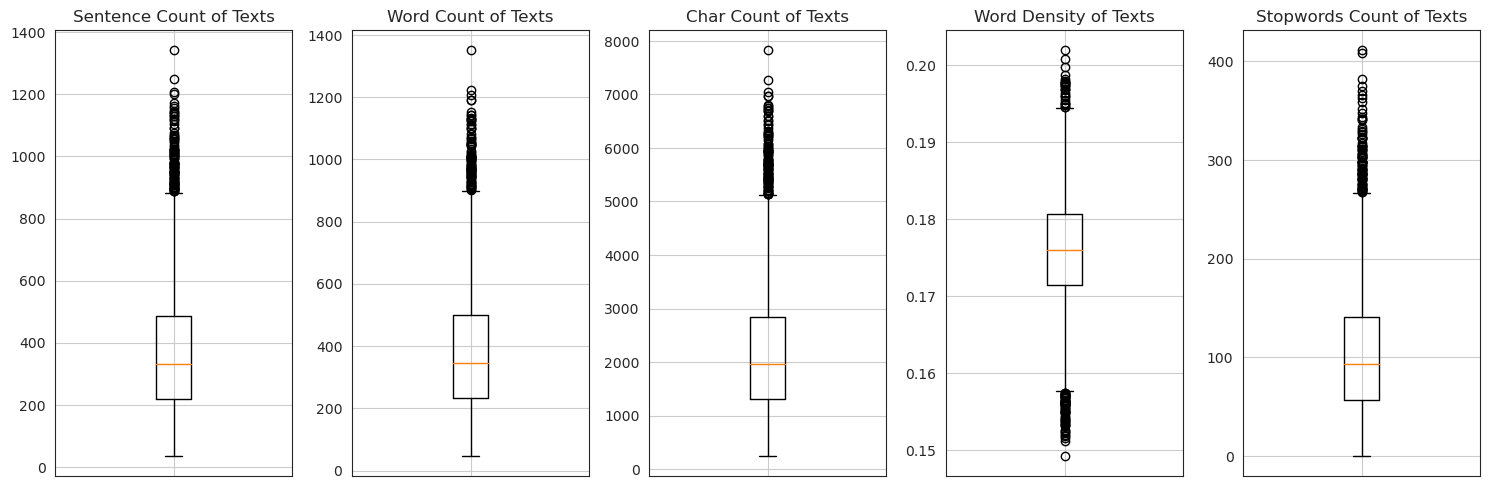}
        \caption*{(a)}
        \label{fig:box_plot_words}
    \end{subfigure}
    \begin{subfigure}
        \centering
        \includegraphics[width=1\linewidth]{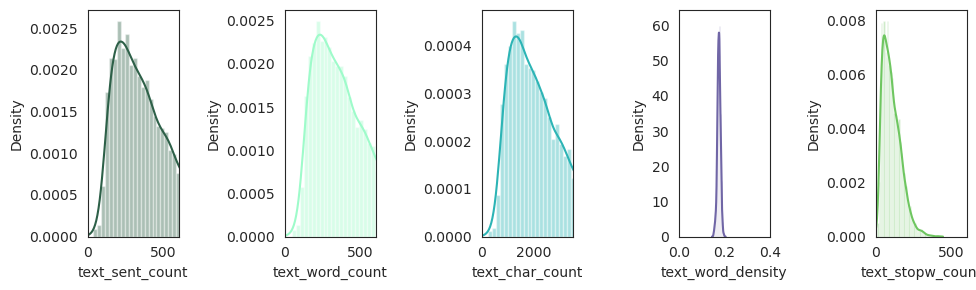}
        \caption*{(b)}
        \label{fig:word_histogram}
    \end{subfigure}
    \caption{
        Illustration of 
        (a) Box plots representing the count of sentences, words,
stop words, and characters in the forecast report;
        (b) Histograms representing the distribution of text in the labels.
        }
    \label{fig:main}
\end{figure}

\begin{itemize}

    \item Count and density distribution of characters, words, and stop words: Figure~\ref{fig:main} represents the count of sentences, words, stop words, and characters in the forecast reports. Also, it illustrates the histograms representing the distribution of text in the labels. 

    \begin{figure} [H]
        \centering
        \includegraphics[width=1\linewidth]{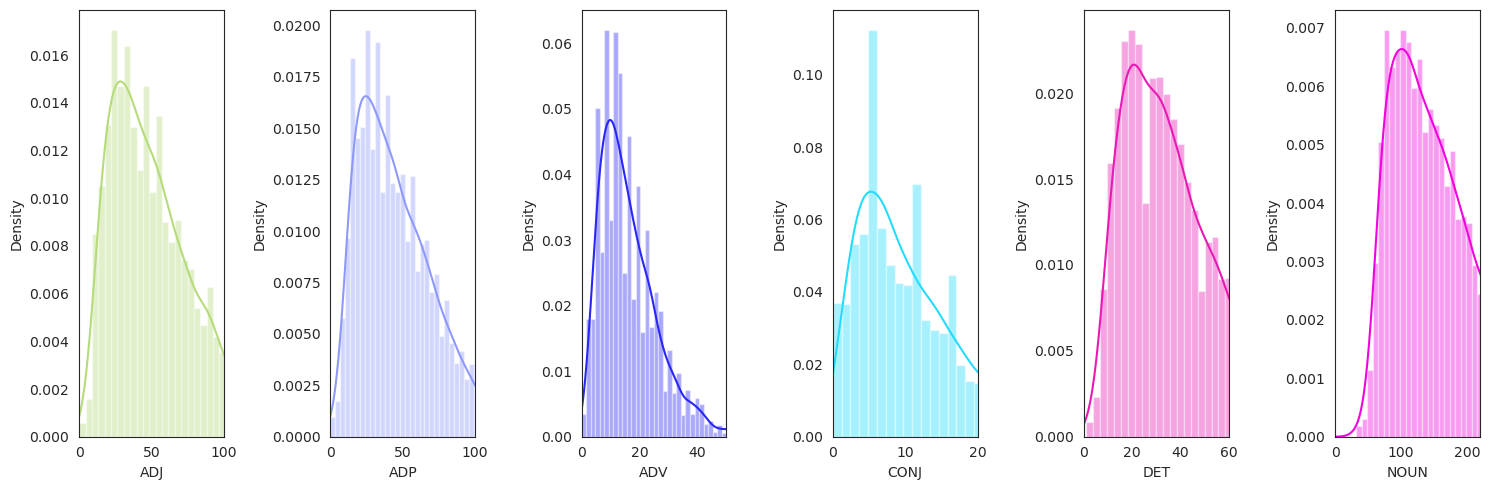}
        \caption{Density distribution of words with respect to part of speech (POS). \footnotesize(\textit{ADJ: adjective, ADP: adpositions, ADV: adverb, CONJ, conjunctions, DET: determiners and NOUN: nouns.})}
        \label{fig:pos_tagging}
    \end{figure}
    
    \item Grammatical categorization and POS tagging: Grammatical or POS tagging maps a word with a corresponding part of speech (nouns, adjective, conjunction, etc.) and plays a key role in understanding the distribution of words in different POS. Figure~\ref{fig:pos_tagging} presents the density distribution of words with different POS tags.     
\end{itemize}

\begin{itemize}
    \item Analysis of top-10 stop words.

    \begin{figure}
        \centering
        \includegraphics[width=0.6\linewidth]{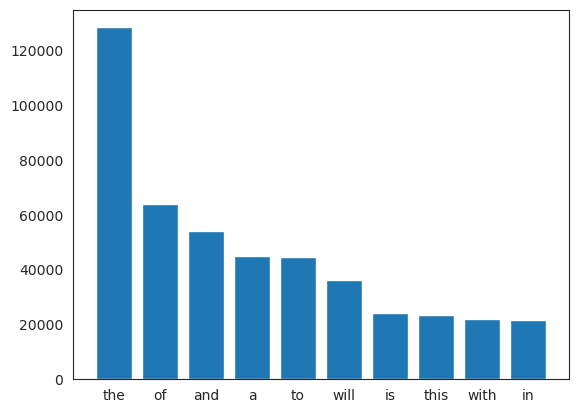}
        \caption{Illustration of histogram for top-10 stop words from the forecast reports.}
        \label{fig:stop_words_a}
    \end{figure}
\end{itemize}

\subsubsection*{Training Details}
\label{app:training}

\begin{figure}[H]
    \centering
    \includegraphics[width=0.75\linewidth]{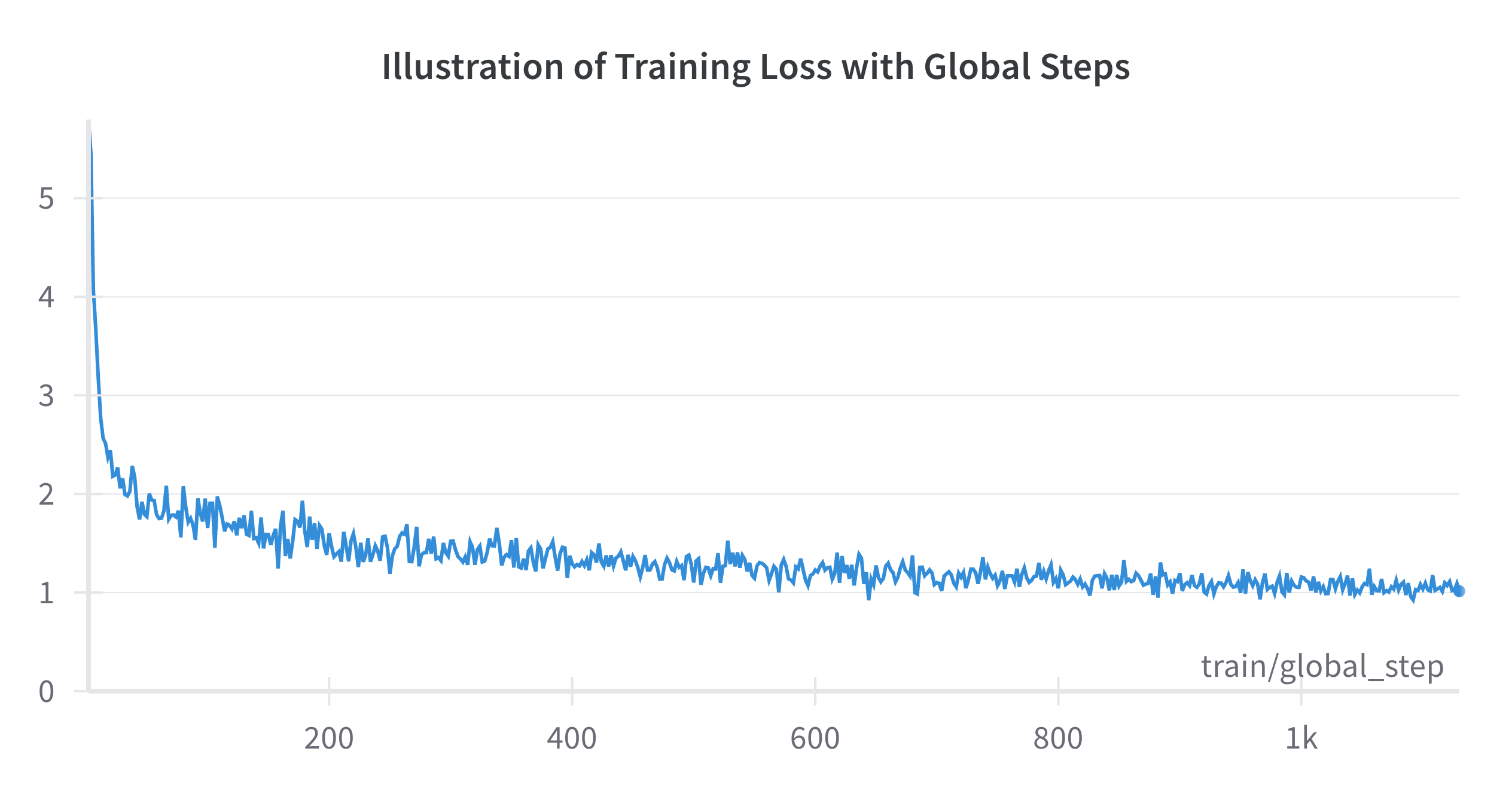}
    \caption{Illustration of the loss curve with the number of steps for finetuning the encoder-decoder captioning architecture.}
    \label{fig:wx-forecast-loss}
\end{figure}

Figure~\ref{fig:wx-forecast-loss} shows the loss curve for fine-tuning the captioning model with the number of steps.

\end{document}